\documentclass{article} 

\usepackage{microtype}
\usepackage{hyperref}
\usepackage{url}
\usepackage{booktabs}
\usepackage{url}
\usepackage[T1]{fontenc}

\usepackage{booktabs}
\usepackage{subcaption}
\usepackage{graphicx}
\usepackage{tcolorbox}
\usepackage{multirow}
\usepackage{float}
\usepackage{calc}
\usepackage{graphicx}
\usepackage{xcolor} 

\usepackage{listings}
\usepackage[usenames,dvipsnames]{xcolor}

\definecolor{deepblue}{rgb}{0,0,0.5}
\definecolor{deepred}{rgb}{0.6,0,0}
\definecolor{deepgreen}{rgb}{0,0.5,0}
\DeclareFixedFont{\ttb}{T1}{txtt}{bx}{n}{8} 
\DeclareFixedFont{\ttm}{T1}{txtt}{m}{n}{8}  

\usepackage[font=small]{caption}

\newcommand\pythonstyle{
\lstset{
language=Python,
basicstyle=\ttm,
morekeywords={self,def},
keywordstyle=\ttb\color{deepblue},
emph={MyClass,__init__},
emphstyle=\ttb\color{deepred},
stringstyle=\color{deepgreen},
frame=tb,
showstringspaces=false,
breaklines=true,                  
breakatwhitespace=true,           
columns=flexible,                 
postbreak=\mbox{\textcolor{deepred}{$\hookrightarrow$}\space}, 
}}
\lstnewenvironment{python}[1][]
{
\pythonstyle
\lstset{#1}
}
{}
\colorlet{punct}{red!60!black}
\definecolor{background}{HTML}{EEEEEE}
\definecolor{delim}{RGB}{20,105,176}
\colorlet{numb}{magenta!60!black}

\lstdefinelanguage{json}{
  basicstyle=\normalfont\ttfamily\small,
  numbers=left,
  numberstyle=\scriptsize\color{gray},
  stepnumber=1,
  numbersep=8pt,
  showstringspaces=false,
  breaklines=true,
  frame=lines,
  backgroundcolor=\color{background},
  literate=
   *{0}{{{\color{numb}0}}}{1}
    {1}{{{\color{numb}1}}}{1}
    {2}{{{\color{numb}2}}}{1}
    {3}{{{\color{numb}3}}}{1}
    {4}{{{\color{numb}4}}}{1}
    {5}{{{\color{numb}5}}}{1}
    {6}{{{\color{numb}6}}}{1}
    {7}{{{\color{numb}7}}}{1}
    {8}{{{\color{numb}8}}}{1}
    {9}{{{\color{numb}9}}}{1}
    {:}{{{\color{punct}{:}}}}{1}
    {,}{{{\color{punct}{,}}}}{1}
    {\{}{{{\color{delim}{\{}}}}{1}
    {\}}{{{\color{delim}{\}}}}}{1}
    {[}{{{\color{delim}{[}}}}{1}
    {]}{{{\color{delim}{]}}}}{1},
}

\usepackage{lineno}

\definecolor{darkblue}{rgb}{0, 0, 0.5}
\hypersetup{colorlinks=true, citecolor=darkblue, linkcolor=darkblue, urlcolor=darkblue}

\title{Beyond the Rosetta Stone: Unification Forces in Generalization Dynamics}

\author{Carter Blum\thanks{Equal contribution.
    \hspace{0.6em}\textsuperscript{$\dagger$}Also at Tel Aviv University.
    \hspace{0.6em}\textsuperscript{$\ddagger$}Also at Harvard University.
    \hspace{0.6em}\textsuperscript{\S}Also at NYU.},
  Katja Filippova\footnotemark[1]{} \& Ann Yuan\footnotemark[1] \\
{\bf Asma Ghandeharioun, Julian Zimmert, Fred Zhang \& Jessica Hoffmann} \\
{\bf Mor Geva\textsuperscript{$\dagger$}, Martin Wattenberg\textsuperscript{$\ddagger$},
  Tal Linzen\textsuperscript{\S} \& Lucas Dixon} \\
Google DeepMind \\
\texttt{\{carterblum, katjaf, annyuan\}@google.com}
}

\usepackage[final]{colm2026_conference}
\begin{document}

\ifcolmsubmission
\linenumbers
\fi

\maketitle
\begin{abstract}
Large language models (LLMs) struggle with cross-lingual knowledge transfer: they sometimes hallucinate when asked in one language about facts expressed in a different language during training.
This work introduces a controlled setting to study the causes and training dynamics of this phenomenon by training small Transformer models from scratch on synthetic multilingual datasets.
Depending on (1) the correlation between facts and the language they were learned in (informativeness), and (2) the ease of language identification (extractability), models either develop unified representations across languages or separate representations; only when representations are unified do facts transfer across languages.
Based on these insights, we propose a unifying perspective which explains a range of prior observations concerning cross-lingual transfer in multilingual LLMs. 
Our work shows controlled settings can shed light on pre-training dynamics and suggests methods to encourage representational unification as part of training that would improve LLMs' cross-lingual transfer.

\end{abstract}

\section{Introduction}

Language models are known to hallucinate facts, especially during cross-lingual factual recall---that is, when facts are learned in one language, but recalled in another language \citep{goldman-eclektic-2025}. 
In this work, we are interested in understanding what gives rise to these cross-lingual hallucinations and how to mitigate them. Therefore, we focus our analysis on the learning dynamics during pretraining \citep{biderman2026position}, and particularly on the impact of  training data properties and the resulting model's representations.

We investigate the connections between data properties, models' representations after learning from this data, and final downstream cross-lingual performance. To do so, we need to 1) have tight control on the training data properties (we want to be able to vary each parameter individually to identify the relevant ones, and causally intervene on some of them to test hypotheses) and 2) analyze many models trained on these different datasets, and the subsequent internal representations. For these reasons, we use a ``Petri dish'' methodology: we train small transformers from scratch on synthetic datasets while systematically and methodically varying their distributional properties.
Such synthetic setups and training data interventions are well-established in the literature, and has already enabled crucial insights into model capabilities \citep{chan-2022-data-dist-properties,park-2024-competition,allen-zhu-physics-2024-tutorial}.

Thanks to this setup, we obtain a plethora of models which have perfect\footnote{Here, 'perfect factual recall' indicates that the in-language test accuracy consistently reaches >99\% after convergence, as demonstrated in Figure\ref{fig:id_vs_ood_test} and the training curves in Appendix Figure \ref{fig:idtest_performance}.} factual recall about facts expressed and recovered in only one language, but a wide range of cross-lingual factual recall accuracy (from 40\% to 100\%). We consequently identify a pivotal early phase where a model develops either unified or separate representations for identical facts across languages. A fully-pretrained model has good cross-lingual factual recall if and only if it develops early unified representations. To capture this effect, we introduce a new metric, the language unification score, and we show that its value in these early steps of training is predictive of cross-lingual generalization in the fully trained model. This score is computed over training examples, and is as predictive of cross-lingual model performance as a validation set would be, but without the need for a held-out set.

We then examine the properties of the data that lead to separate representations, and therefore to subpar cross-lingual generalization. Intuitively, the representations will separate if we learn how to distinguish the languages before we learn the facts. This is more likely to happen in two cases: if the language is \textbf{informative}, that is, if knowing the language helps with learning the facts (e.g. there are more Spanish cities in texts in Spanish), and if the language is \textbf{extractable}, meaning if it is easy to recognize from which language each token/sentence comes (e.g. in Greek, which has a unique script). 

\begin{figure}
\centering\includegraphics[width=0.9\linewidth]{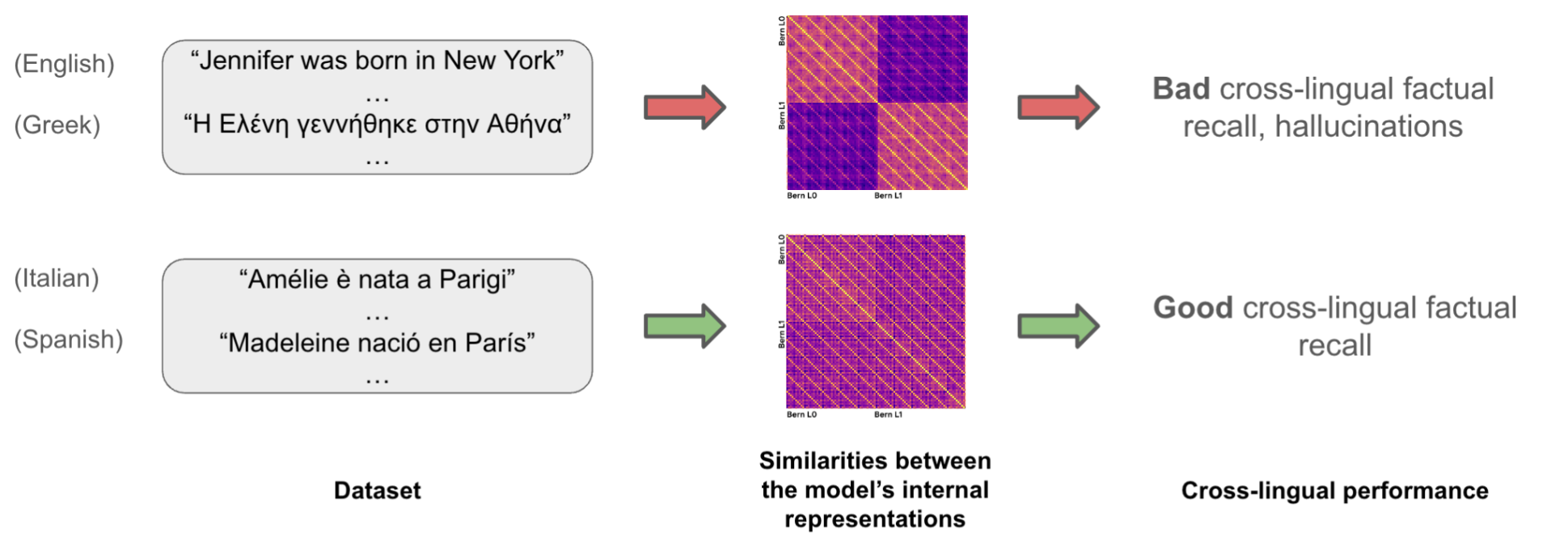}
    \caption{Learning dynamics for cross-lingual factual recall. On top, the languages are informative\protect\footnotemark---aka predictive of content (here, the city)---and extractable---Greek and English are very different. This leads the model to learn separate representation for each language, which then results in poor cross-lingual downstream performance. Inversely, at the bottom, the languages are not as informative (city not correlated with language) nor extractable (Italian is close to Spanish). This leads the model to learn unified representations, which then results in good cross-lingual factual recall.}
    \label{fig:hero}
\end{figure}
\footnotetext{From top to bottom, aside from the first example, the sentences translate to (Greek) "Elini was born in Athens", (Italian) "Amélie was born in Paris", (Spanish) "Madeleine was born in Paris".}
If we increase the amount of parallel data (facts expressed in both languages), we reduce the informativeness of both languages, which in turn increases generalization---a well-known fact. Similarly, our findings provide a unifying perspective on multiple other seemingly disparate observations from prior work about cross-lingual transfer in LLMs.
In summary, our core contributions are: 

\begin{enumerate}
\item We systematically vary parameters to generate synthetic data in \textbf{synthetic languages} with different distributional properties. We train models on each of them and compare in-language and cross-lingual factual recall. By inspecting their internal representations, we identify a crucial phenomenon: the models perform well if and only if they develop \textbf{unified representations} across languages early in training.
\item We show we can measure this unification via the \textbf{language unification score}. This score is computable during early training, and is strongly predictive of downstream cross-lingual generalization. It is as effective for picking the best models as using a validation set.
\item We identify the leading causes of representation unification failure: whether the languages are \textbf{informative} (predictive of content) and \textbf{extractable} (easily differentiable from one another). We then test this hypothesis by artificially increasing and decreasing how informative and extractable the synthetic languages are and thus confirm our findings. This dual lens provides a unified explanation of how script, vocabulary size and embeddings have been shown in seemingly disparate previous works to affect cross-lingual knowledge transfer.

\end{enumerate}

\section{Methodology}\label{sec:methodology}


\paragraph{Data creation and training/test sets} In our tiny setting, we consider only 4 types of facts: A \textit{person} is born in a \textit{city} (birth-city), a \textit{person} is born during a \textit{year} (birth-year), a \textit{person} died in a \textit{city} (death-city), and a \textit{person} died during a \textit{year} (death-year). During our experiments, we use \textbf{synthetic languages}. However, for clarity, the examples in the paper are in English or Spanish. For instance, \texttt{Alice Brown} being born in \texttt{Berlin} is a fact of type birth-city. We create thousands of such facts, and we express them in templates. The potential templates are the same for each fact, and they are arbitrarily split into two sets representing each "language".  By default, the templates do not have overlapping tokens across the two distinct languages; however, multiple templates within the same language may share tokens. The fact arguments (e.g., \textit{persons}, \textit{cities}, \textit{years)}) remain shared across both languages. Our datasets are composed of these facts expressed in these templates, e.g. "\texttt{Alice Brown} was born in \texttt{Berlin}" or "\texttt{Alice Brown}'s life started in \texttt{Berlin}" or "\texttt{Alice Brown} nació en  \texttt{Berlin}" are all valid pieces of data. For each fact, we decide if it is parallel or unparallel data. If it is parallel data, we express this fact in all the templates of both languages, and this data becomes part of the training set. For unparallel data, we pick only one of the two languages, and then select some templates (leaving at least 1 out, similarly to \cite{allen-zhu-physics-2024-tutorial}). The fact expressed in all the templates selected (and therefore in only one language) joins the training set, while the left-out templates in the same language join the \textbf{in-language} test set. The same fact expressed in all the templates of the other language become part of the \textbf{cross-language} test set. In addition to the overall in- (or cross-) language accuracy, for some experiments we report accuracy per fact type. For more details about the generation procedure and the exact templates, see Appendix \ref{app:data-gen}.

Intuitively, increasing the amount of parallel data should improve generalization across languages. To study different regimes, we vary the proportion of cross-lingual events and generate multiple datasets for different ratios, all containing the same facts. 

\paragraph{Tiny model training and evaluation}\label{sec:model}

We train small Transformer models using standard configurations (Pythia, Gemma 2) from random initialization on our synthetic datasets for multiple epochs, monitoring the in- and cross-language accuracies during training. The total parameter count in our models is typically around 2M (six layers, four attention heads, and a hidden size of 128).
As described previously, we vary the following parameters when creating a training set: (1) amount of parallel data, and (2) the discrepancy in entity frequency between languages. We report the fraction of parallel data by event, from 0\% (no parallel data) up to 30\% with a 2\% step size. See App.~\ref{app:training-details} for more details.

\begin{figure*}[t]
  \centering
  \begin{subfigure}{0.45\textwidth}
    \centering
    \includegraphics[width=0.88\linewidth]{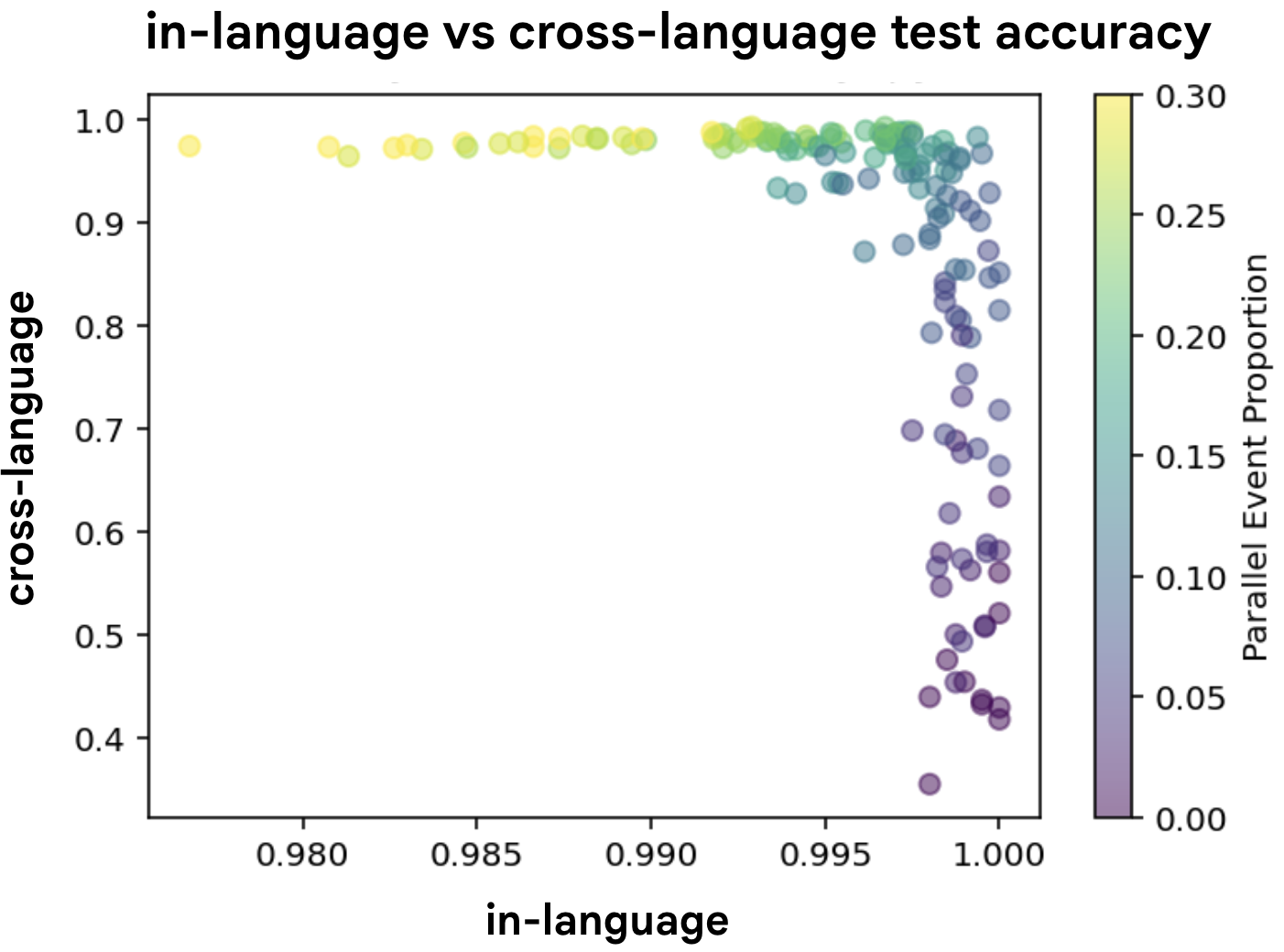}
    \caption{Model family accuracies}
    \label{fig:id_vs_ood_test}
  \end{subfigure}
  \hfill
  \begin{subfigure}{0.45\textwidth}
    \centering
    \includegraphics[width=0.9\linewidth]{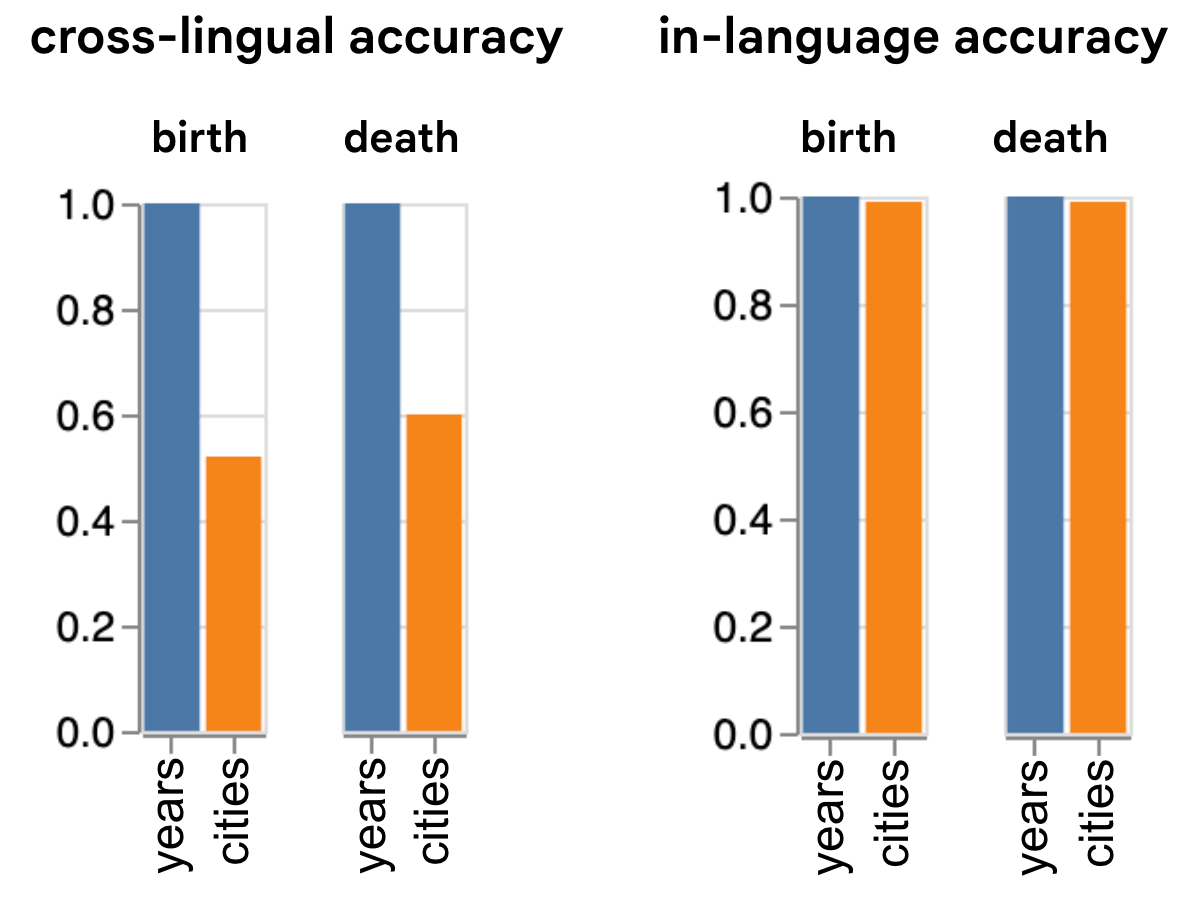}
    \caption{Per-fact-type accuracy}
    \label{fig:id_ood_curves}
  \end{subfigure}

  \caption{\textbf{Left}: In-language vs. cross-language test label probability across Pythia models (pretrained on datasets expressing the same facts) across parallel data ratios. In-language performance does not predict cross-lingual performance. \textbf{Right}: Accuracies for in-language and cross-language evaluation of one model for the four fact types. Model attains perfect cross-lingual accuracy when predicting years but not cities, despite perfect in-language accuracy for that task.}
  \label{fig:ex-eval}
  \vspace{-1em}
\end{figure*}

\section{Learning Stages}\label{sec:dynamics}

\begin{figure*}[t]
  \centering
  \begin{subfigure}{0.32\textwidth}
    \centering
    \includegraphics[width=0.85\linewidth]{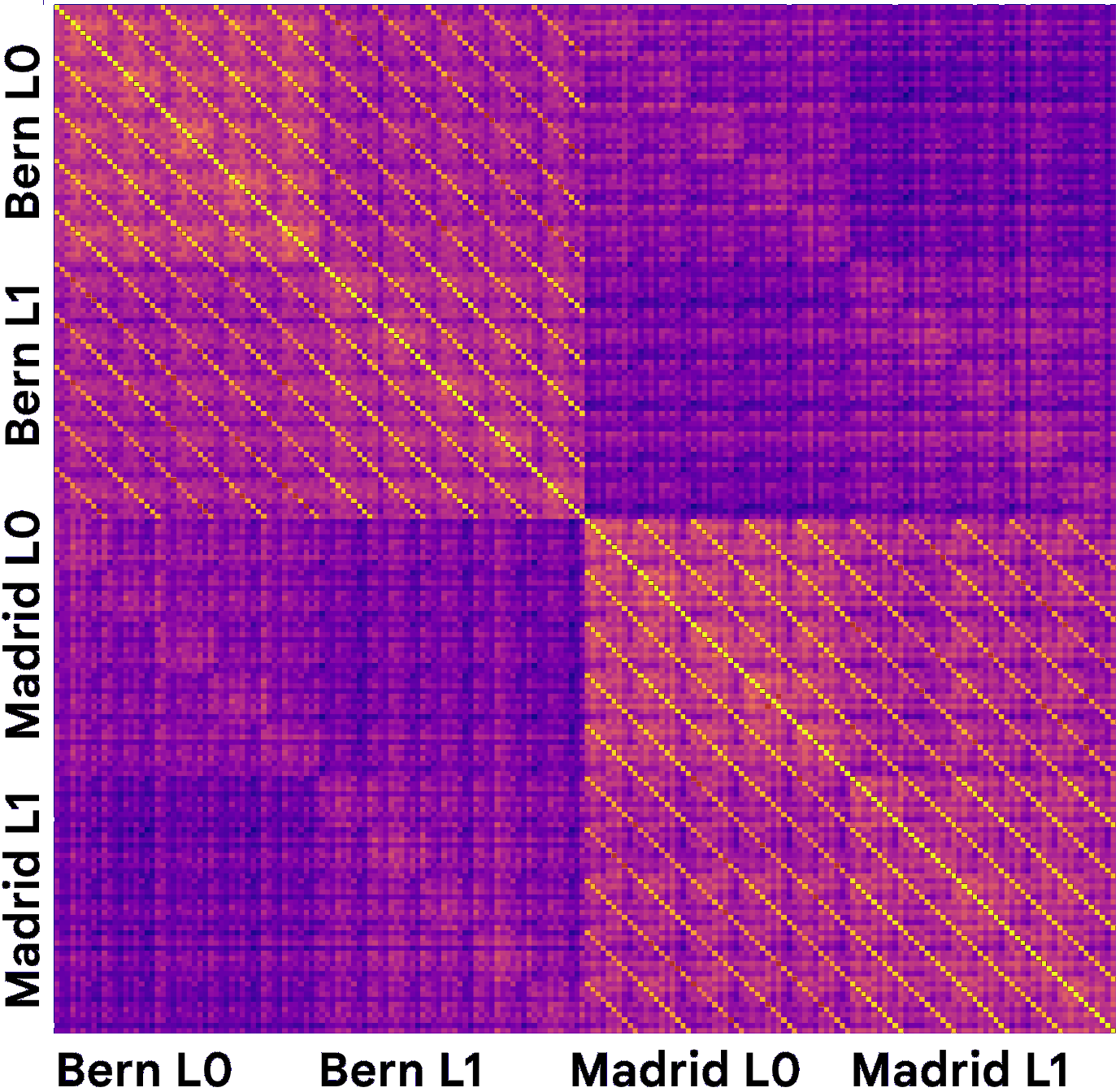}
    \caption{\textbf{Multiple attributes}}
    \label{fig:simple-plaid:median}
  \end{subfigure}
  \hfill
  \begin{subfigure}{0.32\textwidth}
    \centering
    \includegraphics[width=0.85\linewidth]{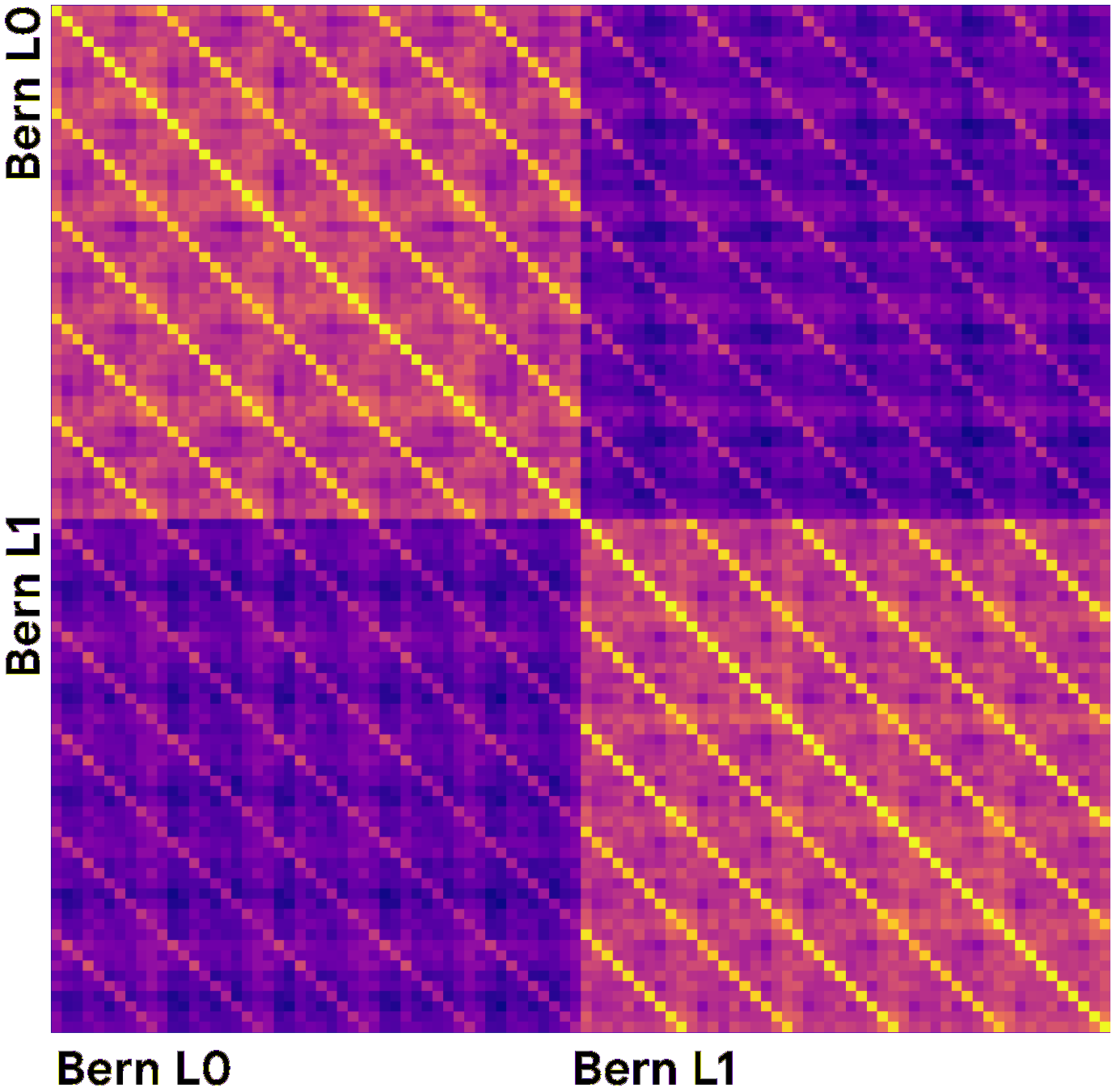}
    \caption{\textbf{Poor} generalization}
    \label{fig:simple-plaid:bad}
  \end{subfigure}
  \hfill
  \begin{subfigure}{0.32\textwidth}
    \centering
    \includegraphics[width=0.85\linewidth]{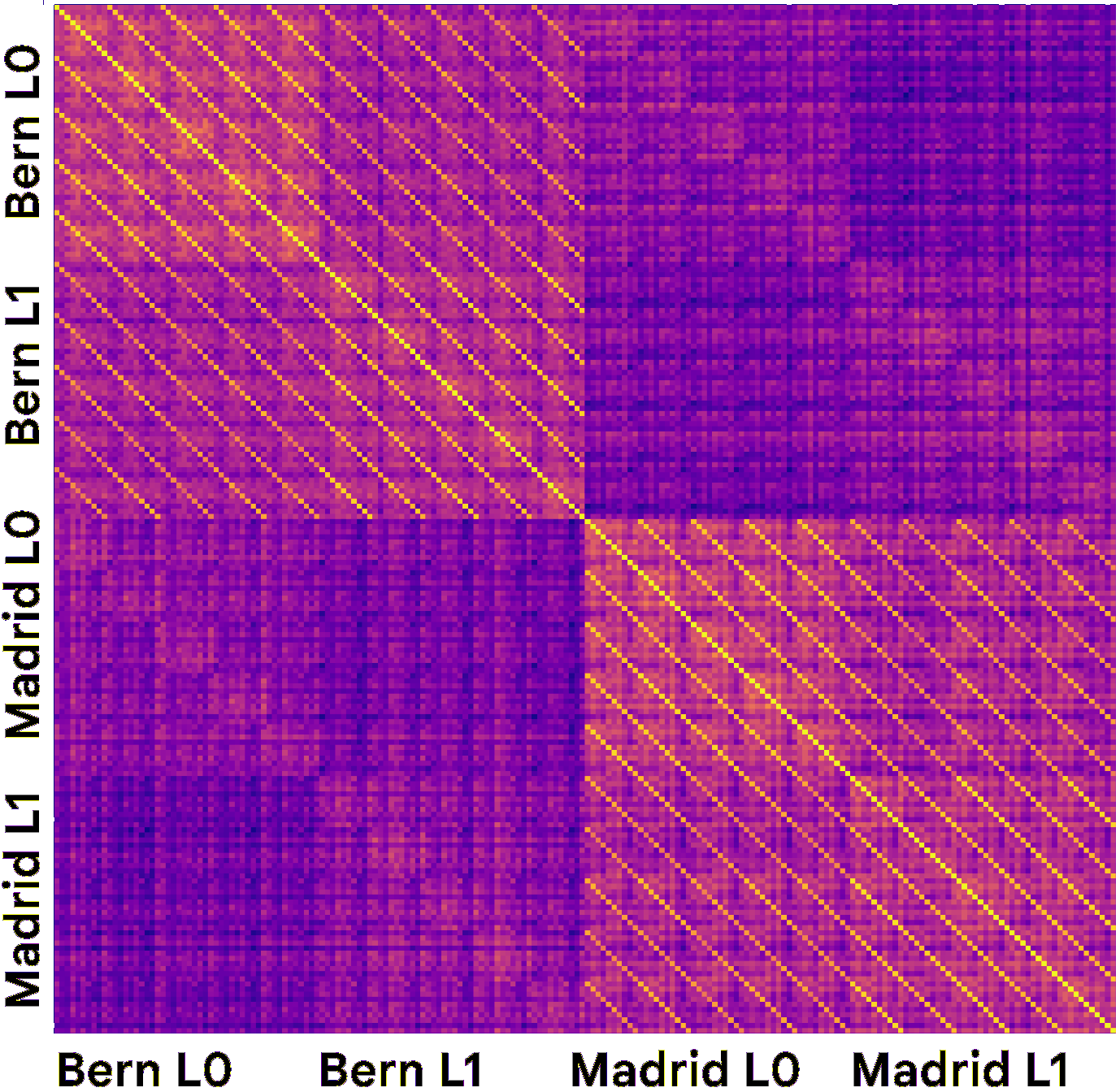}
    \caption{\textbf{Good} generalization}
    \label{fig:simple-plaid:good}
  \end{subfigure}
  \caption{Generalization patterns across different experimental setups.} 
  \label{fig:simple-plaid-main}
\end{figure*}

\begin{figure*}[t]
  \centering
  \begin{subfigure}{0.48\textwidth}
    \centering
    \centerline{[Checkpoint-282]}
    \includegraphics[width=0.96\linewidth]{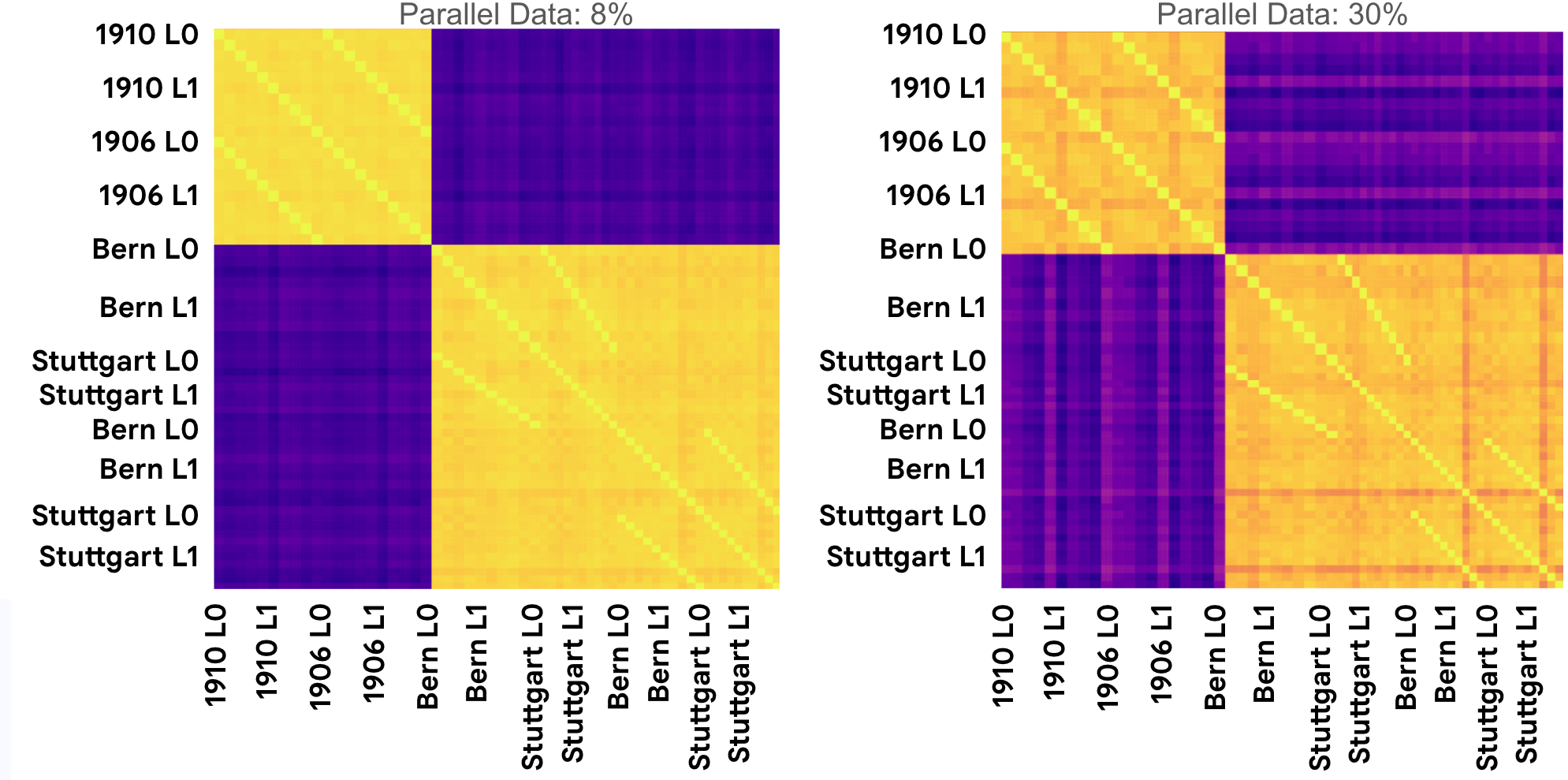}
    \caption{Attribute types (e.g., \texttt{city}) are unified.}
    \label{fig:plaids-ex:c30-ckpt256}
  \end{subfigure}
  \hfill
  \begin{subfigure}{0.48\textwidth}
    \centering
    \centerline{[Checkpoint-564]}
    \includegraphics[width=0.96\linewidth]{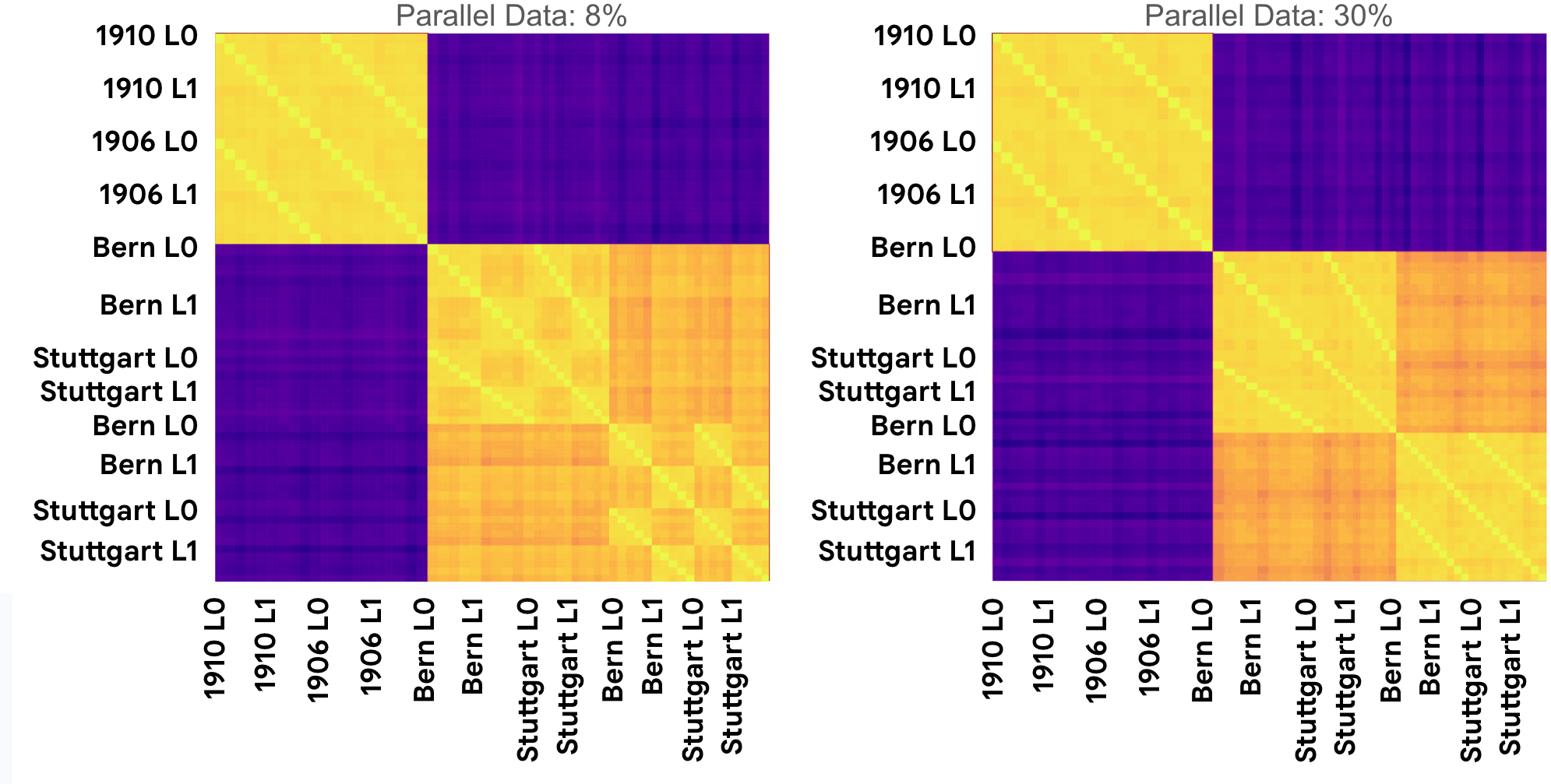}
    \caption{Checkerboard pattern emerges (left).}
    \label{fig:plaids-ex:c30-ckpt512}
  \end{subfigure}

  \vspace{1.5em} 

  \begin{subfigure}{0.48\textwidth}
    \centering
    \centerline{[Checkpoint-1,100]}
    \includegraphics[width=0.96\linewidth]{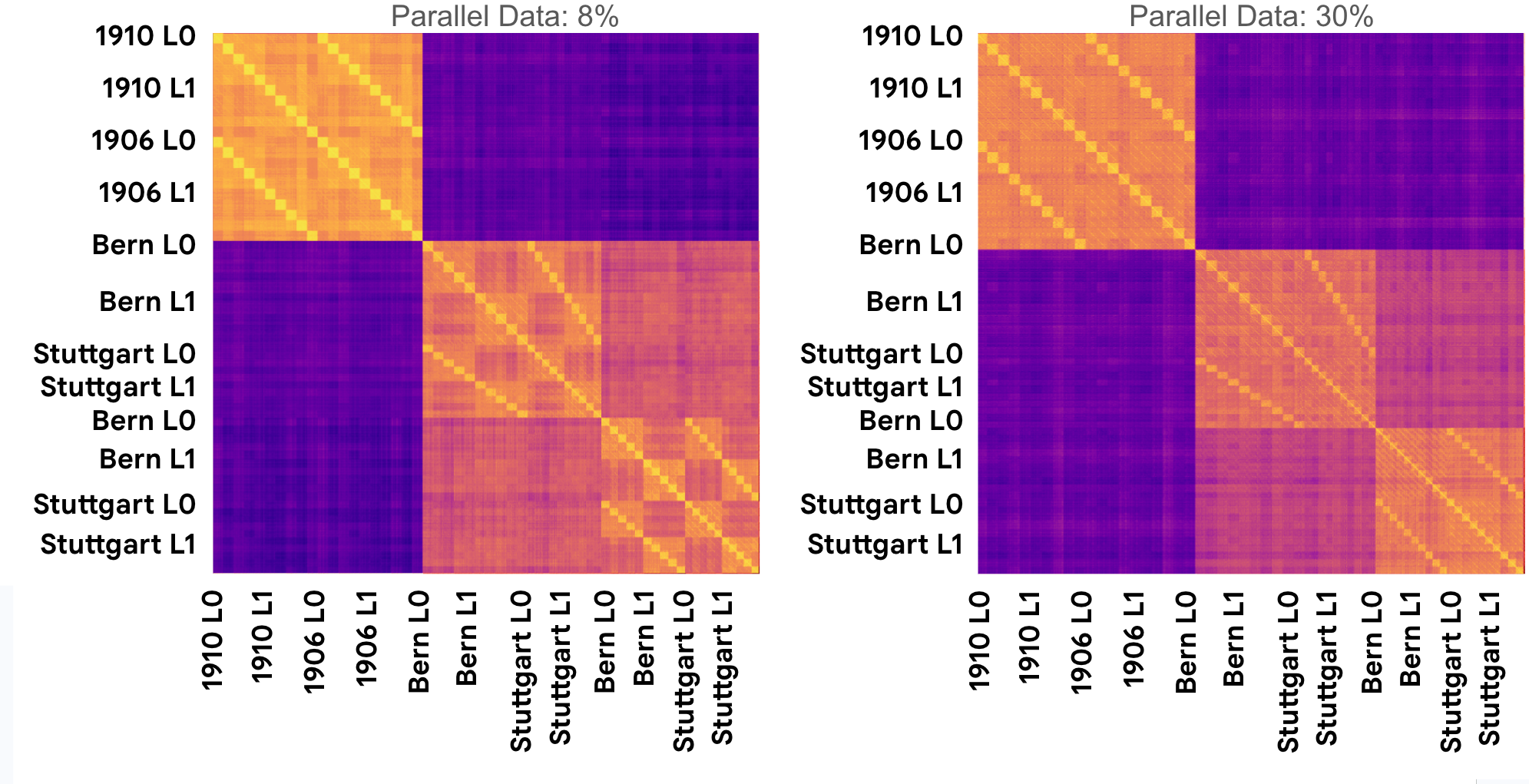}
    \caption{Checkerboarding intensifies (left).}
  \end{subfigure}
  \hfill
  \begin{subfigure}{0.48\textwidth}
    \centering
    \centerline{[Checkpoint-14,000]}
    \includegraphics[width=0.96\linewidth]{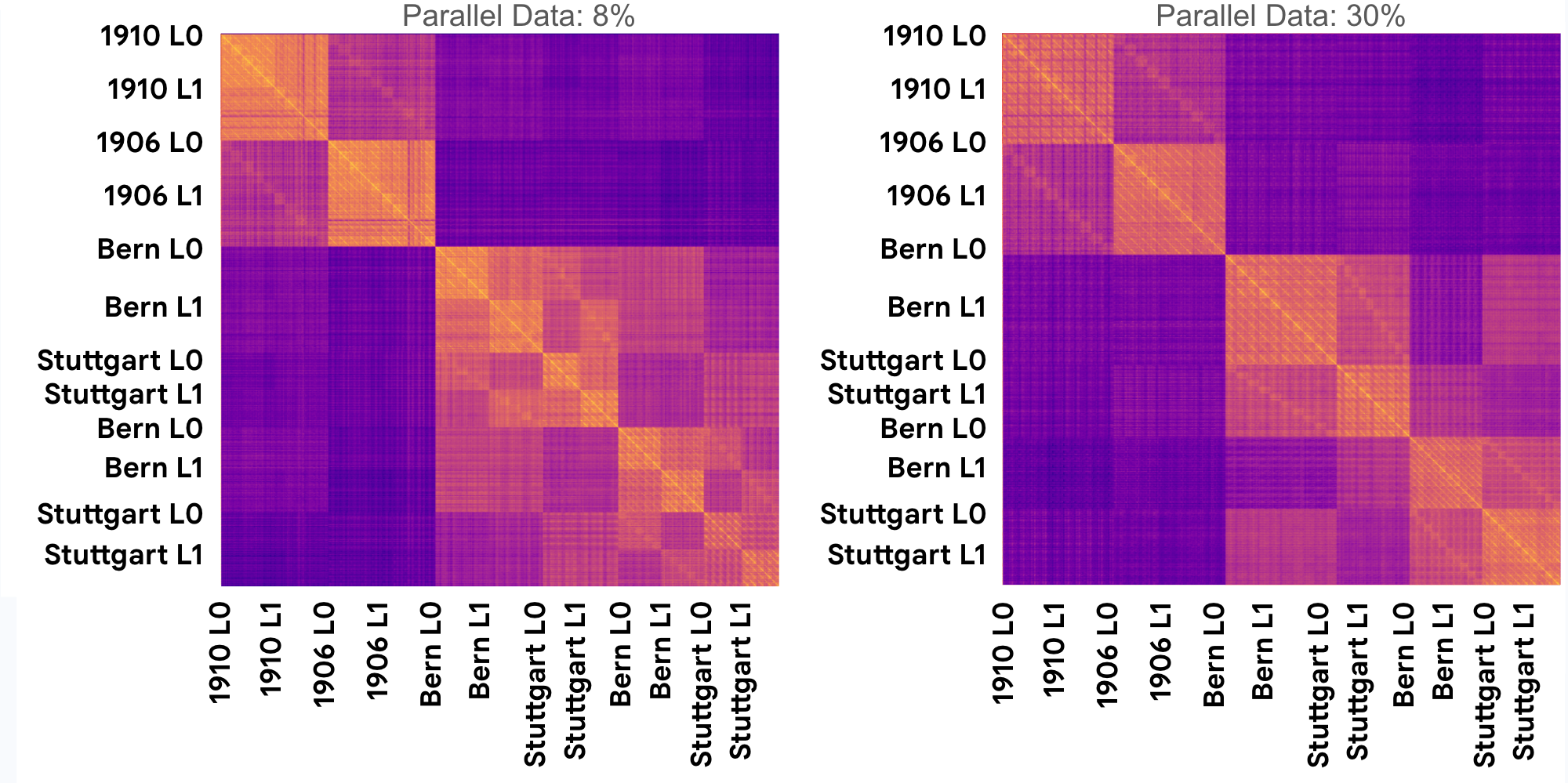}
    \caption{Successful model unifies values (right).}
    \label{fig:plaids-ex:c30-ckpt16k}
  \end{subfigure}

  \caption{Pairwise similarity matrices between activation-based representations at the token preceding the attribute across checkpoints. One model (on the right in every pair of matrices) eventually achieves 100\% cross-lingual test accuracy while the other (on the left) does not. Red means high similarity. Detailed descriptions for each stage are provided in the main text. Note: Labels such as 'Bern' and 'Stuttgart' are repeated on axes to demonstrate the model's clustering of distinct fact types (e.g., birth versus death) for the same city.}
  \label{fig:plaids-example}
  \vspace{-1em}
\end{figure*}

We analyze representations based on: (1) activations from the residual stream, and (2) gradients of model parameters. By monitoring cosine similarity between representations during training, we can pinpoint when semantically related inputs across languages converge and diverge.
To obtain activation-based representations we take the residual stream contents for the token immediately preceding the attribute to be predicted (e.g. \textit{Alice Brown was born \textbf{in}}). 
Unless otherwise noted, we concatenate the activations of each layer to form this representation. The use of such embeddings for the purpose of model analysis and visualization is well established \citep{logit-lens, ghandeharioun-2024}.
To obtain gradient-based representations we use the model weights' gradients at the attribute token. Intuitively, if the gradients between two examples are similar, then they exert a similar influence during training, and are processed by similar model parameters. 
We draw inspiration from methods for identifying influential training inputs by comparing the gradients of training data to test data \citep{koh-liang-if-2017, schioppa-if-theory-2024, ruis-2025-procedural}, and leverage a recent, computationally efficient approximation developed by \citet{chang-scalable-if-2024}.

Fig.~\ref{fig:simple-plaid:median} shows pairwise cosine similarities between activation-based representations\footnote{Gradient-based plots are in Fig.~\ref{fig:app:plaids-ex:gradient} in the Appendix.} of a Gemma model trained on a 50-50 language split and 16\% cross-lingual events. In these and following experiments, we adopt a 50-50 language split and a moderate percentage of cross-lingual events to simulate a balanced bilingual environment with a realistic fraction of parallel data. This setup allows us to isolate the effects of language unification without introducing confounding variables like dataset imbalance. Representations are computed from training examples and, while referring to distinct \textit{birth-year} facts, all have \textit{Bern} or \textit{Madrid} as the (correctly predicted) birth-city attribute. Within each attribute, examples are grouped by language (\textit{L0, L1}). In this model with median generalization ability, examples with the same attribute (e.g., \textit{Madrid}) are more similar to each other than to examples with a different attribute (\textit{Bern})---this is clearly visible in the two large red blocks. However, within the red matrix corresponding to the same attribute (e.g., \textit{Madrid-Madrid}, bottom right), two sub-blocks are visible, indicating that examples of the same language are more similar than examples across languages. For comparison, Fig.~\ref{fig:simple-plaid:bad}-\ref{fig:simple-plaid:good} show the similarities for a \textit{single} birth-city attribute (\textit{Bern)}, again grouped by language, from two models---with worse and better cross-lingual factual recall (8\% vs.~30\% cross-lingual events). Only Fig.~\ref{fig:simple-plaid:bad} has four distinct blocks, indicating cross-lingual dissimilarity.

When does separation by language happen in training? Consider Fig.~\ref{fig:plaids-example}, which shows pairwise similarity matrices for two models across checkpoints. One model (on the right in every pair of matrices) eventually achieves perfect cross-lingual transfer while the other (on the left) does not. The models' training sets are identical with respect to factual content, languages used, language split (50-50) and number of examples, but differ in the amount of parallel data---30\% events vs.~8\%. Note that the trends on display are observed consistently across dozens of models trained with different languages, language proportions, model scales, etc.
Each matrix in Fig.~\ref{fig:plaids-example} shows pairwise similarities between 300 examples equally split between \texttt{birth-year, birth-city, death-city}, each corresponding to two values (\textit{1906, 1910} for \texttt{birth-year}, \textit{Bern, Stuttgart} for \texttt{birth-city}, \textit{Bern, Stuttgart} for \texttt{death-city}). Again, within each attribute value, examples are grouped by language (\textit{L0} and \textit{L1}). We observe that poorly-generalizing models undergo a signature phase (around checkpoint-564) wherein language identity, rather than semantic equivalence, drives representational similarity.

We quantify our observations with a \textit{language unification score}, which predicts cross-lingual generalization. Intuitively, the language unification score measures how much the model \textit{avoids} the checkerboarding on the right side of Fig. ~\ref{fig:plaids-example}.

\section{Unification Predicts Cross-Lingual Performance}\label{sec:unification}

\begin{figure}[h]
    \centering
    \includegraphics[width=0.7\linewidth]{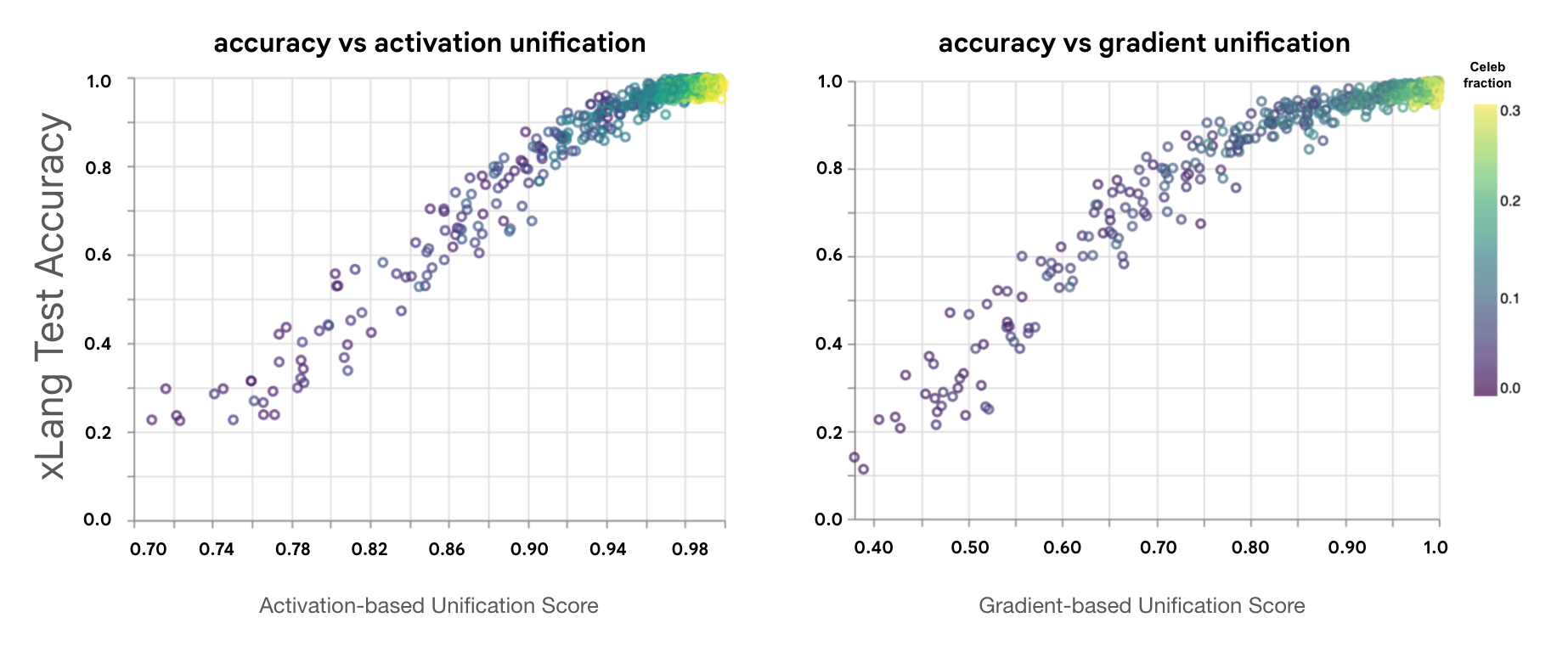}
    \caption{Activation language unification scores (left) and gradient language unification scores (right) correlate strongly with cross-lingual factual recall accuracy (0.97 and 0.94 PCC, respectively) (see Fig.~\ref{fig:activ-ood-layerwise} for layerwise results). Each datapoint corresponds to a different model training run. Note that the fraction of parallel examples (denoted by color) also correlates strongly with generalization ability.}
    \label{fig:activ-ood}
\end{figure}

Concretely, the language unification score captures the similarity between semantically equivalent cross-lingual datapoints against a baseline of similarity between semantically distinct same-language datapoints. We define it as $\mathrm{Unification}(\theta, \mathcal{D}) := E_{X, Y\sim \mathrm{Facts}(\mathcal{D})}\big[sim_\theta(X, Y) / sim_\theta(X, X)\big]$ where $X,Y \in \mathrm{Facts}(\mathcal{D})$ samples the datapoints corresponding to each fact in the dataset $\mathcal{D}$, grouped by language. $X$ and $Y$ are representations of the same fact in different languages (see Appendix Fig.~\ref{fig:unif-score-pseudocode} for pseudocode). For our experiments, we let $\mathcal{D}$ be the parallel examples from the training set. $sim_\theta$ is the average cosine similarity between the representations of both sets.
The language unification score correlates strongly (Pearson's correlation coefficient $>$0.95) with cross-lingual accuracy, for both activation-based and gradient-based representations (Fig \ref{fig:activ-ood}). 

\begin{figure*}[t]
  \centering
  \begin{minipage}{0.48\textwidth}
    \centering
    \includegraphics[width=0.98\linewidth]{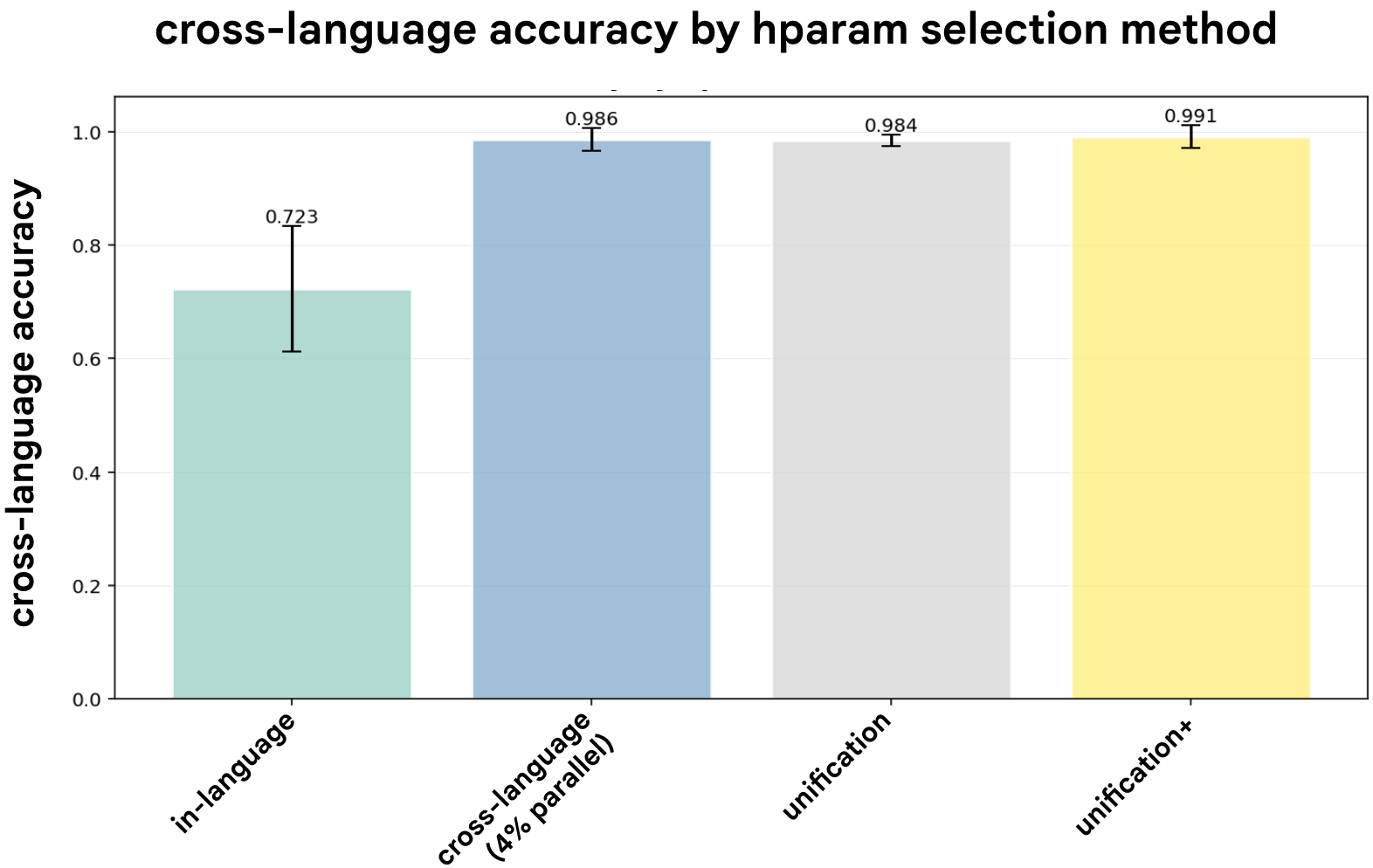}
  \end{minipage}
  \hfill
  \begin{minipage}{0.48\textwidth}
    \centering
    \includegraphics[width=0.9\linewidth]{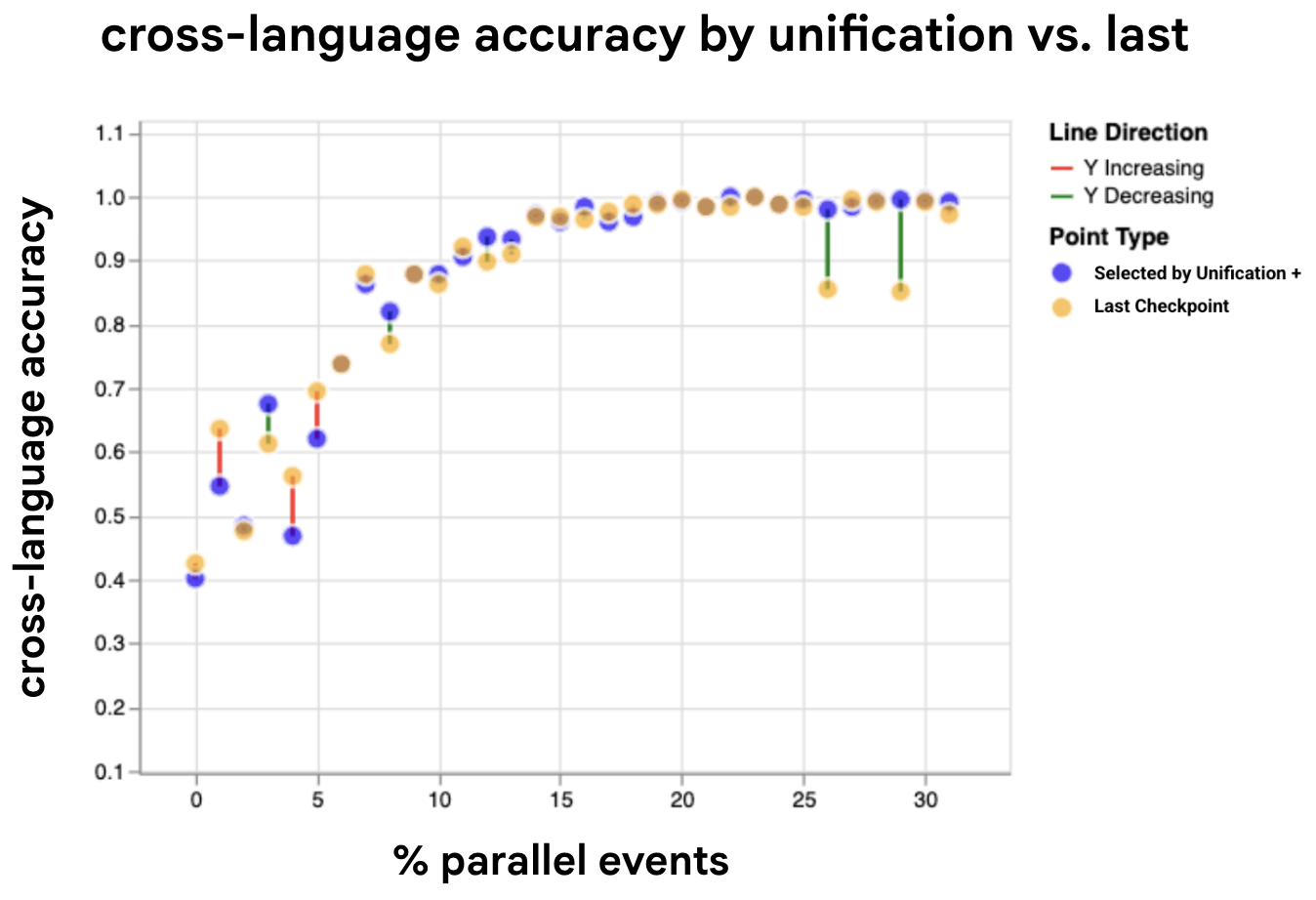}
  \end{minipage}

  \caption{Comparison of hyperparameters and checkpoint selection strategies.} 
  \label{fig:hparams-checkpoint-combined}
  \vspace{-1em}
\end{figure*}

Not only does the language unification score correlate strongly with model quality, it can be used to \textit{select} training runs that will generalize well. In fact, as Fig.~\ref{fig:hparams-checkpoint-combined} (left) demonstrates, we find that utilizing the language unification score can be as effective as collecting a small test set. These experiments were performed over 110 runs varying the fraction of parallel examples (from 0\% to 20\%), the ratio of monolingual examples in the majority vs minority language (from 1:1 to 20:1) and the amount of token overlap within a language. We repeatedly sampled 33\% of runs and compared the cross-lingual test performance of the runs chosen by a variety of heuristics.
These heuristics include:
    \texttt{in-Lang Test} (selecting the model with the best in-language test performance),
    \texttt{xLang Test (k\%)} (selecting the model based on a smaller cross-lingual test set, with k\% of the cross-lingual test set being used for model selection. This is a strong baseline, as it effectively uses more data), and
    \texttt{language unification score} (last token language unification score).

\noindent
We also evaluate language unification scores as a method for selecting a checkpoint, as it may be expensive or difficult to collect cross-lingual validation data. In Fig.~\ref{fig:hparams-checkpoint-combined} (right), we compare to a baseline that selects the last checkpoint among those with the highest in-language accuracy. While both methods tend to choose checkpoints with similar quality, selecting using language unification scores avoids some scenarios where the model begins overfitting on in-language recall. 
This represents an exciting new direction for model selection -- mechanistic analysis of models can improve our ability to predict generalization. 

\section{Analyzing the Language Feature}\label{sec:lang-signal}

As the previous analysis demonstrates, and in line with observations from analyses of LLMs, cross-lingual factual recall fails if the model's internal representations are separated by language, that is, when \textit{language identity is strongly encoded} in the hidden representations.
We hypothesize that the strength of this encoding, or the language feature footprint, is determined by the \textbf{informativeness} of language identity for the prediction task and its \textbf{extractability} from the data.
This is theoretically grounded: seminal work by \citet{saxe-2019-mathematical} shows that models learn features in descending order of the variance they explain in the training data; e.g., a high-level \textit{plant-vs-animal} distinction is learned before fine-grained categories like \textit{bird-vs-fish}.
Similarly, \citet{lampinen-2024-learned} argue that easier-to-learn features tend to dominate a model's internal representations and explain more variance than difficult-to-extract features. 
To restate our hypothesis: 
\begin{tcolorbox}[colframe=blue!25,
colback=blue!10,
coltitle=blue!20!black, title={}]
\small
If the language identity provides a useful signal about the label (is \textbf{informative}) and is easy to recognize (is \textbf{extractable}), it will be learned early and dominate the representation. Conversely, if language identity is not informative or hard to extract, its representational footprint will be small and representational separation is less likely.
\end{tcolorbox}

\subsection{Language informativeness}\label{sec:language_informativeness}
\vspace{-1em}

\begin{figure}[H]
\centering
\begin{subfigure}[b]{0.49\textwidth}
\centering
    \includegraphics[width=0.98\linewidth]{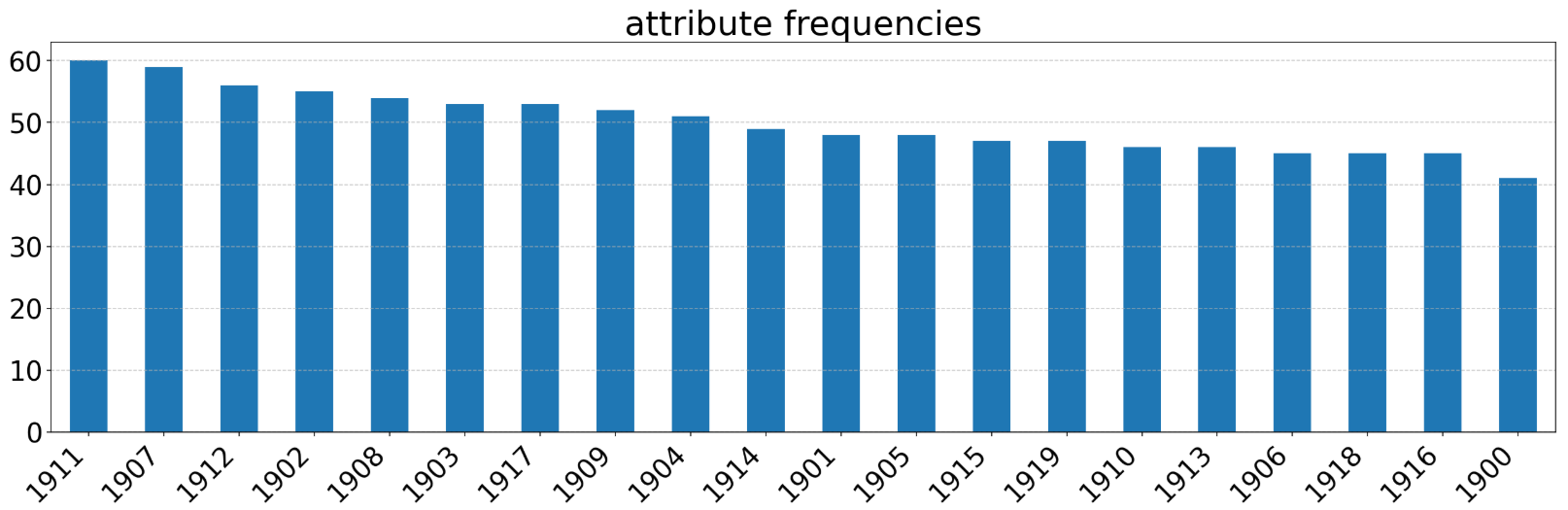}
\end{subfigure}
\hfill
\begin{subfigure}[b]{0.49\textwidth}
\centering
    \includegraphics[width=0.98\linewidth]{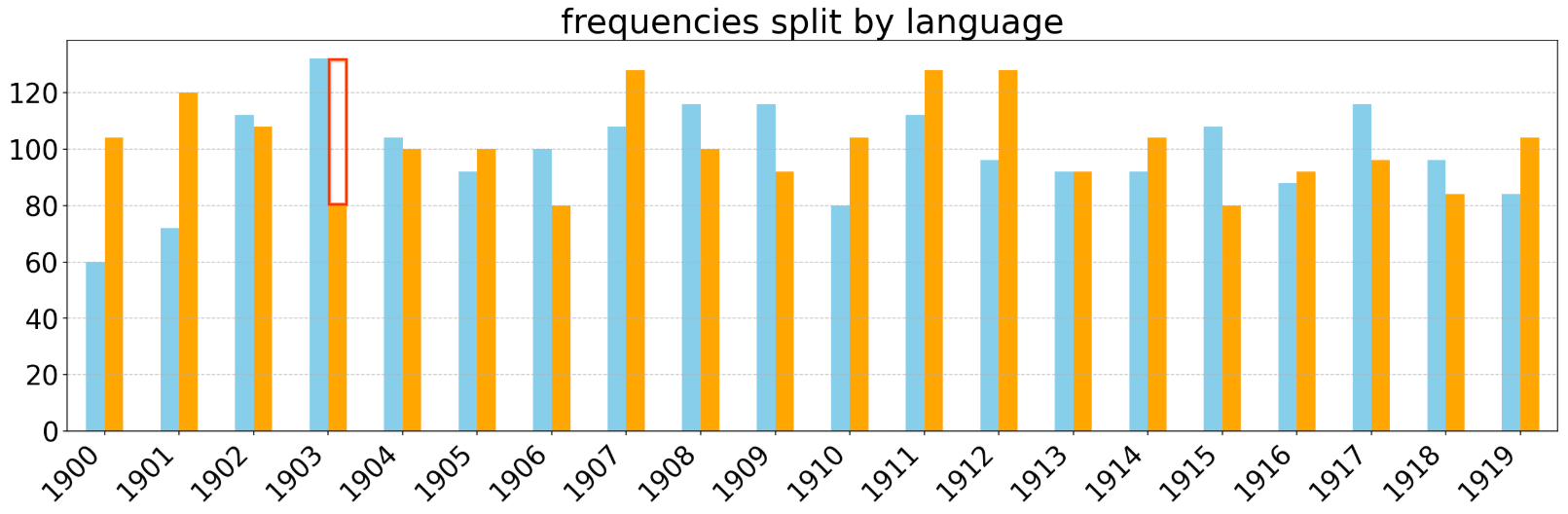}
\end{subfigure}
\caption{Frequencies of the birth-year attribute values in the Knowledge Graph (KG) (left) and in a dataset, split by language (right). Light blue and orange on the right represent the frequencies in the first and second languages respectively after template verbalization. The red rectangle highlights the difference in example counts between two languages for a particular year: it represents the number of examples for that attribute needing to be added to the second language (orange) to create a balanced dataset, versus being added to the first language (blue) to create an imbalanced dataset. }
\label{fig:year-dists}
\vspace{-1em}
\end{figure}

How can the language of an example be a useful signal for prediction? Strictly speaking, attributes (e.g., \textit{1911}) are a function of the subject entity (e.g., \textit{John Smith}) and the fact type (\textit{birth-year}). However, the distribution over attribute values is not uniform in our Knowledge Graph (KG) or datasets (Fig.~\ref{fig:year-dists}). Thus, a spurious correlation (positive correlation) exists between the language identity and the predicted attribute. 
This property of our toy data mimics real multilingual datasets: for example, Spanish texts reference Spanish cities more often than German texts do. The language feature does therefore provide a useful prior over attribute values, helping to reduce loss early in training, \textit{before} the model learns to recall facts by combining the subject and relation representations \citep{geva-dissecting-2023}.
To measure the role of the mutual information between the language and the attribute variables, we create two dataset versions from an existing base dataset with little ($<$10\%) parallel data and with an equal split between languages: \texttt{balanced} includes additional examples (all corresponding to \textbf{new} events) to equalize the example count for every attribute value across the two languages, thereby \textit{minimizing} the language feature's informativeness without increasing the amount of parallel data; and \texttt{imbalanced} includes the same number of additional examples, created from the same set of new events, but adds them in the language in which the attribute is already present more frequently, amplifying the existing discrepancies and therefore \textit{increasing} the language feature's informativeness.

\begin{figure*}[t]
  \centering
  \begin{subfigure}{0.46\textwidth}
    \centering
    \includegraphics[width=0.95\linewidth]{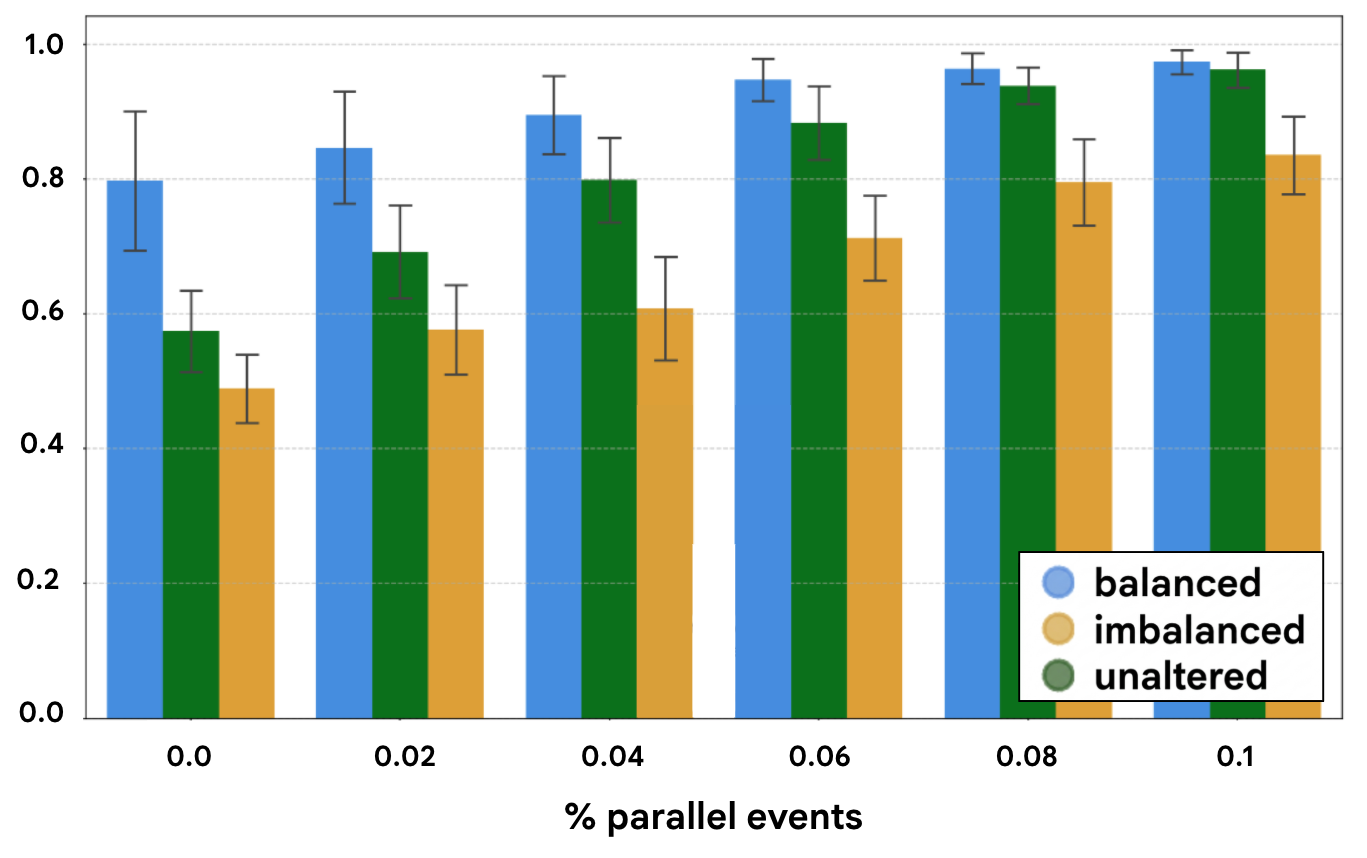}
    \caption{Accuracies by attribute distributions}
  \end{subfigure}
  \hfill
  \begin{subfigure}{0.46\textwidth}
    \centering
    \includegraphics[width=0.95\linewidth]{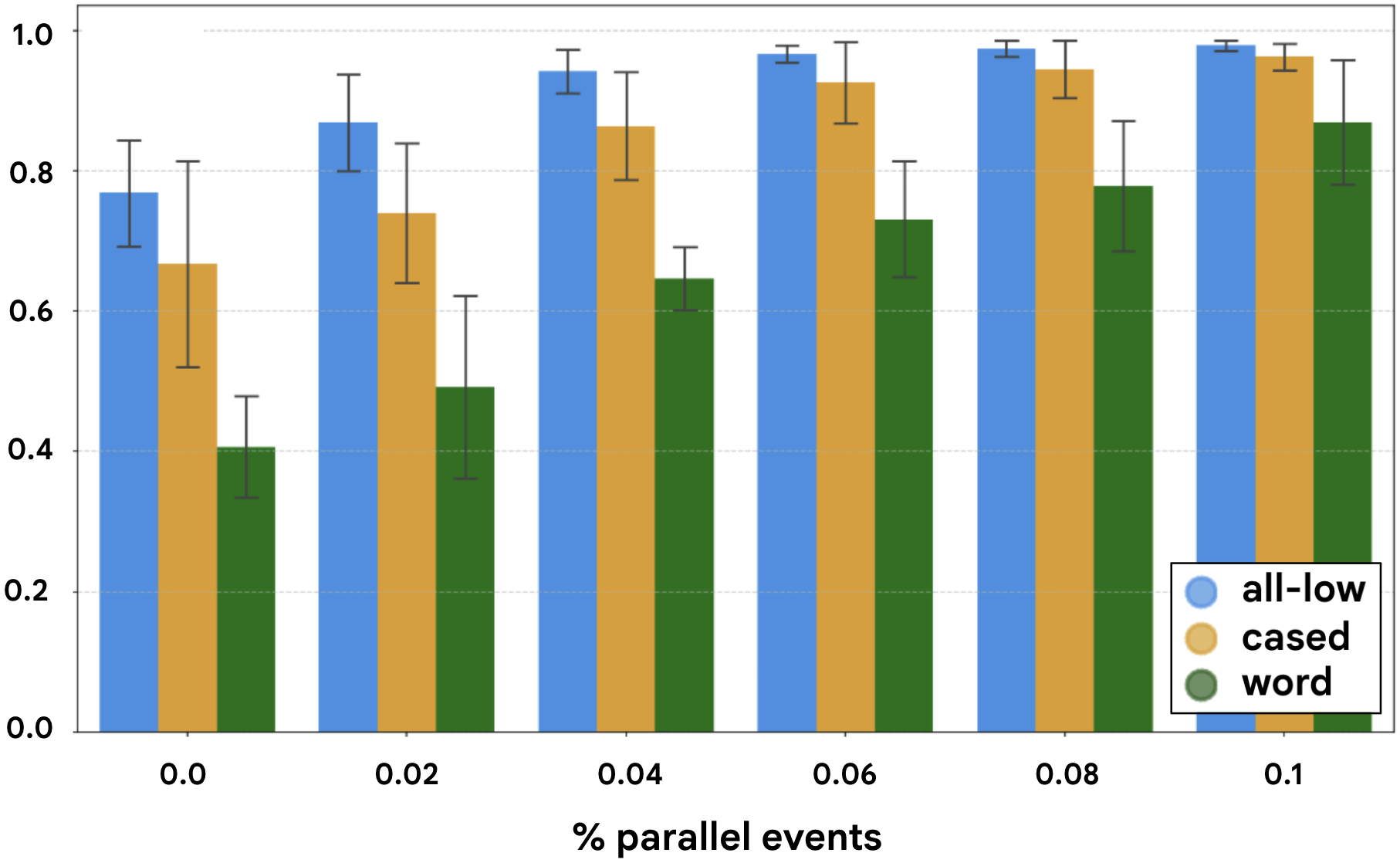}
    \caption{Accuracies for different tokenizers}
  \end{subfigure}

  \caption{Cross-lingual accuracy for different values of parallel events for models (a) trained on unaltered, \texttt{balanced} and \texttt{imbalanced} datasets, (b) trained with the character (\texttt{all-low} and \texttt{cased}) and \texttt{word} tokenizers.}
  \label{fig:results:aligned}
  \vspace{-1em}
\end{figure*}

\begin{figure}[H]
    \centering
    \includegraphics[width=0.45\linewidth]{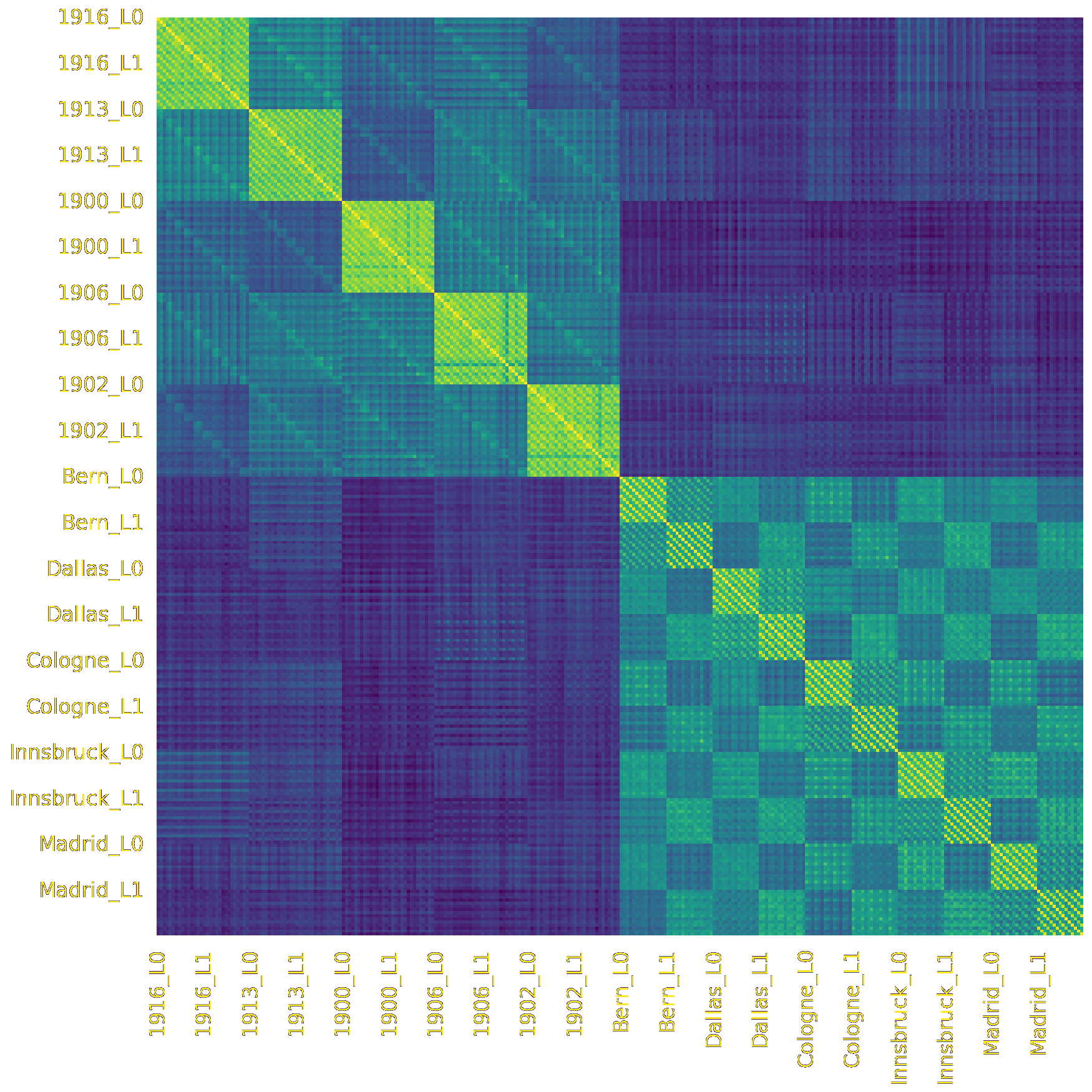}
    \hfill
    \includegraphics[width=0.45\linewidth]{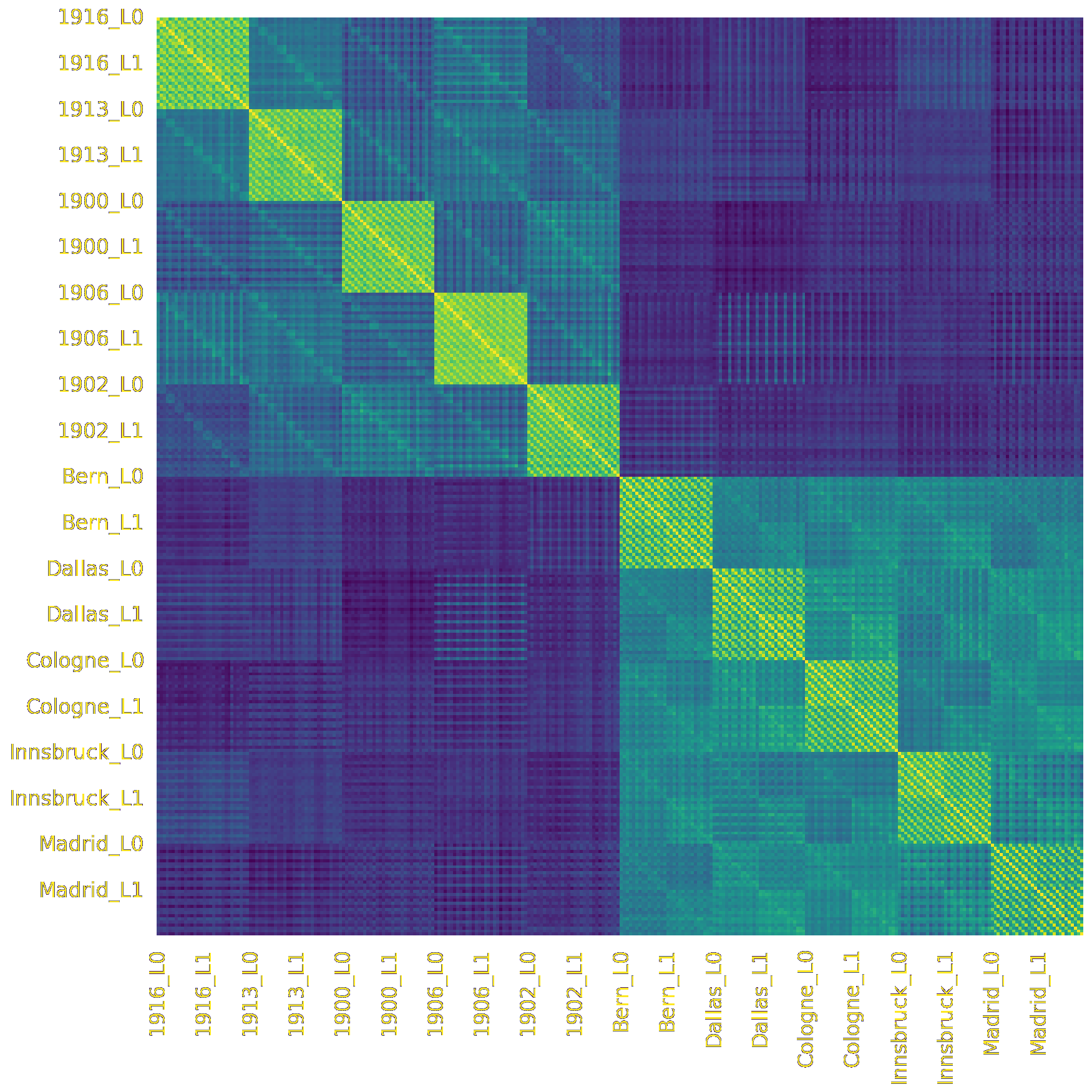}
    \caption{Imbalanced vs.~balanced at checkpoint-40,000}
    \label{fig:app:plaids-ex:misaligned}
\end{figure}

\noindent
Fig.~\ref{fig:results:aligned} (left) shows cross-lingual accuracies for the three settings. Cross-lingual generalization is consistently worst when attribute distributions are imbalanced and correlation between language and attribute is high (checkerboarding is correspondingly more prevalent in this setting, see App.~Fig.~\ref{fig:app:plaids-ex:misaligned}). 
This experiment reveals a clear correlation between the informativeness of the language feature and cross-lingual performance: cross-lingual accuracy improves even when the amount of facts expressed in both languages stays zero. 

\subsection{Language extractability}\label{sec:lang-signal:extractability}

We next test the hypothesis that making the language feature easier to extract also boosts its influence on the model's representations. In our default setup, the language feature is trivial to extract: our synthetic languages do not share vocabulary, and our tokenizer is word-based. Thus every token (except for entity arguments like names, cities, and years) is a perfect indicator of language identity.
To make the language feature harder to extract, we switch to a \textit{character-based} tokenizer while keeping all other aspects of model training unchanged. A character model needs to accumulate information over many tokens to recognize the language. Furthermore, to isolate the effect of language feature extractability from other potential effects of this tokenizer change, we contrast three settings: \texttt{word} Baseline word-based tokenizer and the original dataset (highly extractable language feature); \texttt{all-low} Char-based tokenizer, with all the templates in lowercase (less extractable); \texttt{cased}: Char-based tokenizer, where templates in the first language are in lowercase and the second are uppercase (more  extractable language feature than \texttt{char} but less than \texttt{word} because people's names are spelled normally, introducing lowercase letters).

\begin{figure}[H]
    \centering
    \includegraphics[width=0.45\linewidth]{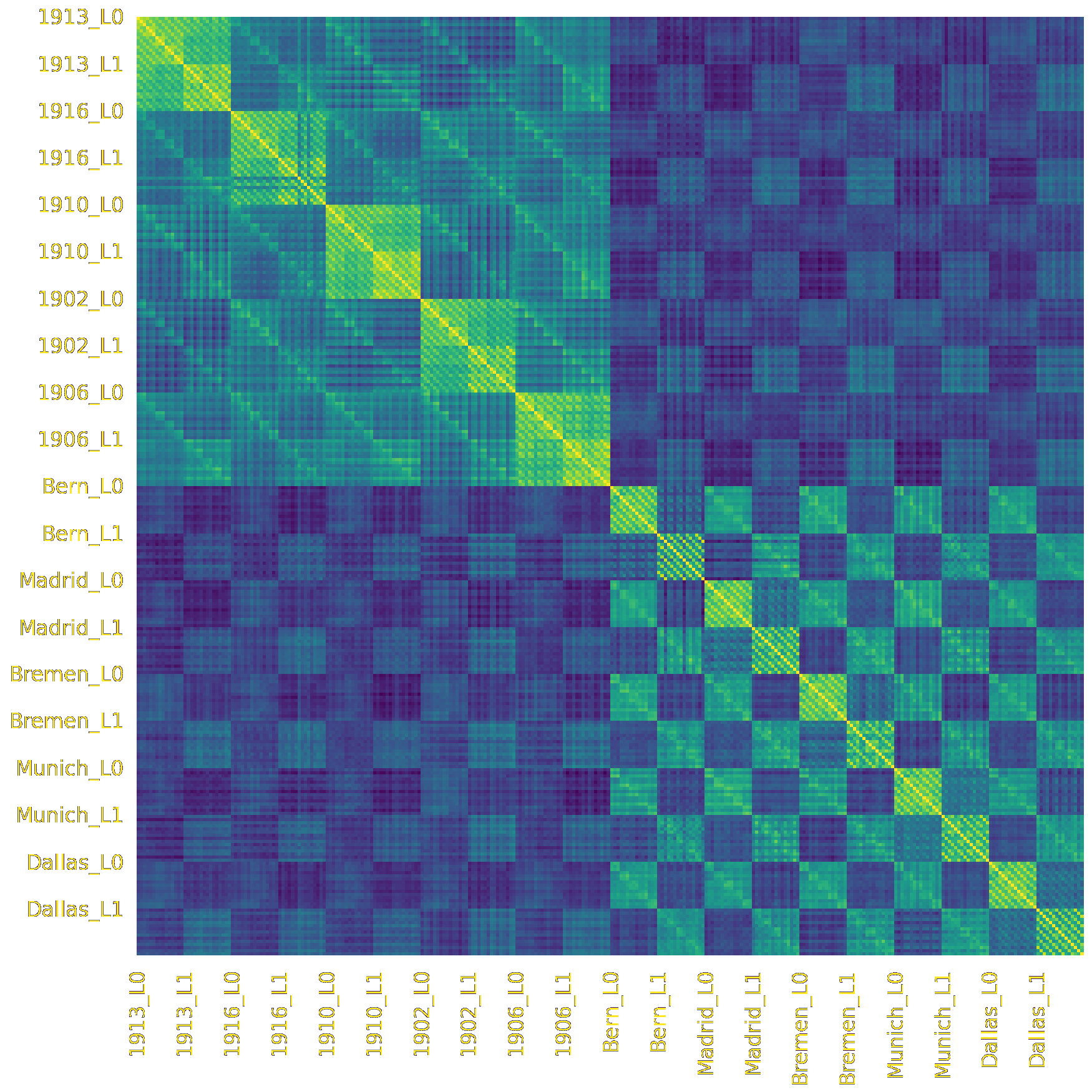}
    \hfill
    \includegraphics[width=0.45\linewidth]{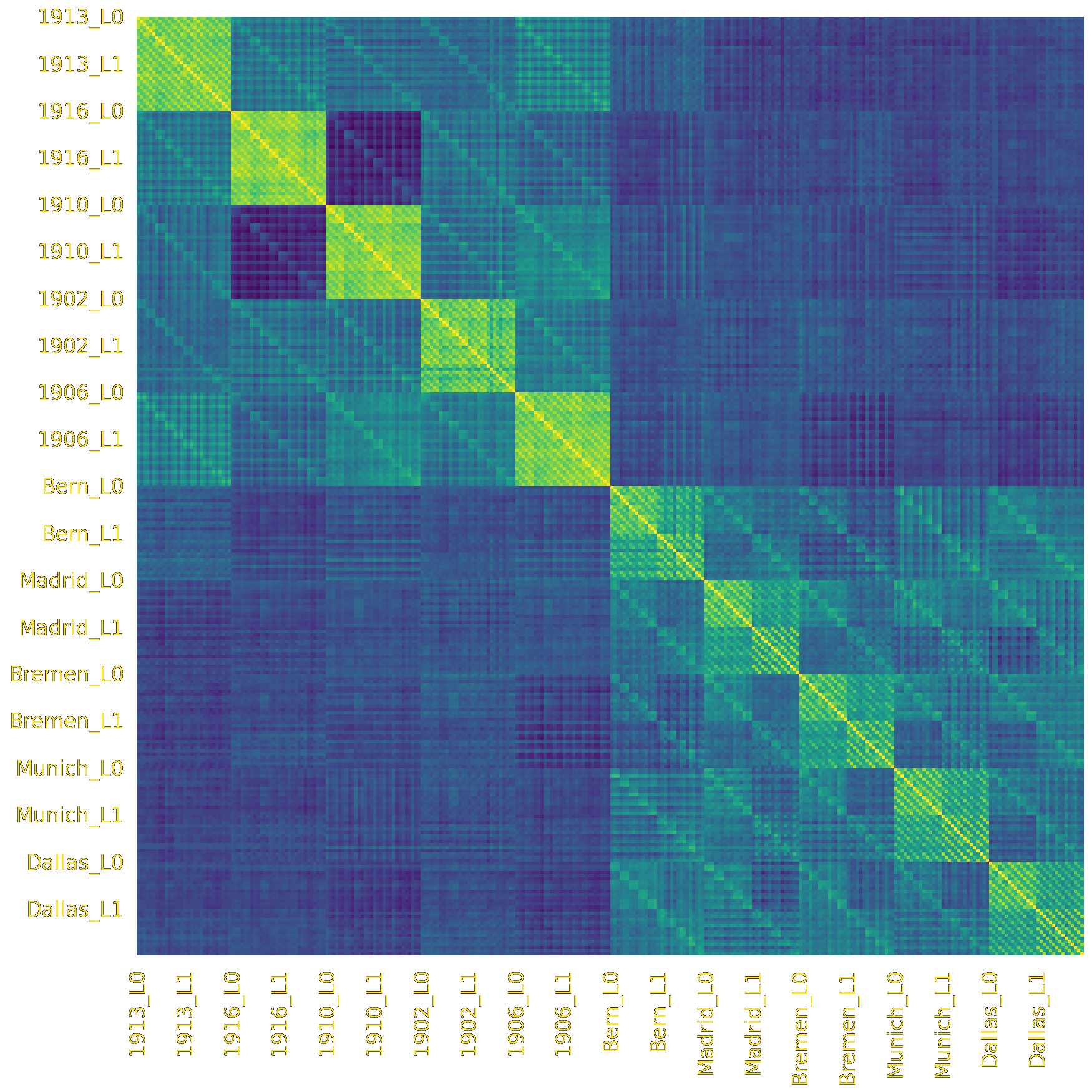}
    \caption{Cased vs all-lowercased (by language) at checkpoint-8,000}
    \label{fig:app:plaids-ex:cased}
  \end{figure}

\noindent
The last two models are directly comparable, as both use the same character-based tokenizer on the same examples. The only difference is that the language feature is more easily extracted by the latter model, since the token sets used in the two languages are disjoint (upper-case vs lower-case characters). 
The results in Fig.~\ref{fig:results:aligned} (right) confirm that when the language feature is more difficult to extract, it has a direct, positive effect on cross-lingual generalization: for the same percentage of parallel data, the \texttt{char-all-low} models, where the language feature is least extractable, consistently achieve the highest accuracy. Similarly, checkerboarding is more present in `cased` models (App.~Fig.~\ref{fig:app:plaids-ex:cased})

We observe a similar effect when making language harder to extract by increasing the number of templates. We conduct experiments where there is no token-level indication of language, making it difficult for the model to correctly group all templates into languages. Increasing the number of templates substantially improves generalization, as demonstrated in App.~\ref{app:increasing-templates}, 
while neither increasing the number of events by 10x nor increasing the training duration by 10x improves generalization.

Recall that Figure \ref{fig:id_vs_ood_test} shows both that parallel data improves cross-lingual generalization, and that generalization can take place without parallel data. 
The experiments in this section suggest a new perspective on parallel data's role in cross-lingual transfer: Increasing parallel data naturally reduces the language feature's informativeness about attribute distributions.

\section{Interpreting Existing Observations}\label{sec:related:explained}

Our key contribution is to shed light on why language models might fail to transfer factual knowledge cross-lingually. Here we show how our findings
help explain various seemingly unrelated observations in prior work:

\paragraph{Script similarity} 
Several studies have observed that cross-lingual performance correlates more strongly with script similarity than with geographical, linguistic, or cultural proximity. For instance, Greek—an Indo-European language—shows low correlation with other European languages \citep{liu-2025-tracing-multilingual}. Conversely, Indonesian—an Austronesian language utilizing the Latin script—demonstrates better cross-lingual transfer with English than other Asian languages \citep{goldman-eclektic-2025}. Further supporting this, research has shown that handling unseen languages \citep{muller2021when} or strengthening low-resource translation \citep{whenscriptsdiverge2025} heavily depends on mitigating script divergence.

While seemingly counter-intuitive, our framework offers a clear explanation: language identity is trivial to extract from a unique script (e.g. Greek). Consequently, the "language feature" is absorbed early in training, leading to siloed representations. In contrast, shared scripts (like Latin) delay this extraction, allowing for better transfer.

\paragraph{Aligning embeddings} 
\citet{prealign} initialize embeddings for aligned words to be similar prior to pretraining to promote shared representations. Rather than directly replicating these interventions, our framework provides a mechanistic rationale for why they work: pre-alignment and increased vocabulary overlap obscure the language feature, reducing its extractability. This encourages the early development of the unified representations that we show are critical to cross-lingual transfer.

\paragraph{Shared vocabulary} 
\citet{patil-2022-overlap} propose increasing vocabulary overlap between languages prior to training, demonstrating that this improves cross-lingual knowledge transfer. These empirical benefits of shared vocabulary were also highlighted in early studies of Multilingual BERT \citep{pires2019how, k2020crosslingual} and large-scale cross-lingual representation learning \citep{conneau2020unsupervised}, which identified subword overlap as a key driver of zero-shot transfer. Following the same logic, \citep{whenscriptsdiverge2025} proposes mapping distinct scripts to shared phonemes, which dramatically improves cross-lingual generalization. We interpret this result similarly: increased vocabulary overlap makes the language identity feature harder to extract, thereby reducing its representational footprint and improving cross-lingual accuracy. 
Furthermore, \citet{qi-etal-2023-cross} find that vocabulary overlap correlates with cross-lingual factual consistency. They thus speculate that larger models do not necessarily show better cross-lingual consistency because their vocabularies are larger. We add another layer of interpretation: larger vocabularies simplify language identification, which in turn increases the prominence of spurious language features. 

\paragraph{Other work} 
Our findings also contextualize other observations in real LLMs. \citet{mousi-etal-2024-exploring} note that deeper layers exhibit more cross-lingual alignment, mirroring our observation that language checkerboarding is most prominent in lower layers. Furthermore, while \citet{wen-yi-mimno-2023-hyperpolyglot} find that models like XLM-R cluster embeddings by script and mT5 by semantic similarity, our framework suggests these representational differences may arise from how distinct training objectives influence the informativeness and extractability of the language feature.

\paragraph{Vocabulary Overlap in More Recent LLMs}

To replicate our observations regarding the effects of tokenization (Sec.~\ref{sec:lang-signal:extractability}), we study vocabulary overlap between languages as a predictor of cross-lingual consistency, following \citet{qi-etal-2023-cross}. 
We re-run the vocabulary similarity experiment by \citet{qi-etal-2023-cross} with \textbf{four recent open LLMs}: Gemma-3-4B\citep{team2025gemma}, Llama-3.2-3B \citep{grattafiori2024llama3herdmodels}, Qwen3-4B \citep{yang2025qwen3}, and Mistral-7B \citep{jiang2023mistral7b}. We use Flores-200 \citep{costa2022no} dataset that consists of 2,000 sentences translated from English into different languages. We run those sentences through the model tokenizer to get a model-specific vocabulary for each language. For each pair of languages $A$ and $B$ with vocabularies $V_A$ and $V_B$, we calculate their Jaccard similarity ($S = |V_A \cap V_B| / |V_A \cup V_B|$). We then group examples by their source and target language, and for a more intuitive analysis, focus on examples with the source language of English. We calculate the Pearson correlation between $S$ and cross-lingual consistency across models (defined here as the consistency metric for factual queries introduced by \citet{qi-etal-2023-cross}) and observe an average coefficient of 0.69, ranging from 0.43 to 0.85 across different models (per-model details are in App.~\ref{app:vocab-sim}). This finding aligns with \citet{qi-etal-2023-cross} which reported a positive correlation between cross-lingual consistency and vocabulary similarity showing that vocabulary overlap explains a significant amount of variance in cross-lingual transfer.

\section{Discussion and Related Work}\label{sec:discussion}

Many studies look into causes of model hallucinations and  attribute them to training and sampling noise, gaps in pretraining data \citep{xu-2024-hallucination-inevitable}, and misaligned incentives in post-training \citep{schulman_rl}. However, these fail to explain \textit{cross-lingual} factual errors, nor does increasing model scale solve the problem \citep{aggarwal-lm-factuality-2025,qi-etal-2023-cross}. 
Prior work finds that LLMs may develop both a lingua franca for factual knowledge (typically based on English) and distinct language silos \citep[inter alia]{aggarwal-lm-factuality-2025,lim-2025-understanding,schut-2025-multilingual, lim-language-latent}, and that their hidden representations can be language-agnostic or language-specific depending on the layer \citep{wang-lost-2025}. This builds upon foundational work demonstrating that shared cross-lingual structures naturally emerge in pretrained models \citep{conneau2020emerging}, and that successful zero-shot transfer requires representations to align across languages before predictive capabilities can transfer \citep{wu2021first}. 
However, the root cause of these phenomena is not understood, as most studies treat models as static artifacts. Such analysis, while valuable, cannot explain how knowledge \textit{arises} during training, and cannot lead to effective pretraining interventions. While some have investigated the training dynamics of knowledge acquisition in multilingual LLMs \citep{zeng-2025-converging-lingua-franca,liu-2025-tracing-multilingual}, their approach is non-interventional and doesn't establish a causal link between data properties and cross-lingual transfer. 

Our work connects and contributes to the body of research into spurious correlations \citep{geirhos-2018-imagenet-trained,mccoy-2019-right-for} which loom behind many surprising model failures and generalization challenges \citep{book-of-why}.  Similar to recent work by \citet{hermann-2020-what-shapes} and \citet{lampinen-2024-learned}, we investigate models’ inductive biases and aim to characterize how a feature’s complexity and predictivity influence its representation. However, we do so in a setup that imitates factual knowledge acquisition and transfer in LLMs, where feature complexity (i.e., low extractability) and predictivity (i.e., informativeness) \textit{emerge naturally} from the standard training process.
In addition, our results demonstrate that the language feature, while obsolete once the true feature is learned, persists in representations (App.~\ref{app:facts-vs-lang-footprint}). This has implications for fine-tuning when the model again may prefer to rely on an easier feature \citep{lovering-2021-predicting}.

Our results suggest two possibilities for improving cross-lingual recall: by obscuring differences between languages, or by balancing attribute frequencies across languages in the pretraining mixture. However language clearly \textit{can} be a useful prior, e.g., in factual queries requiring a language- or culture-specific answer (though how much the model relies on it depends on how easily the language feature can be extracted). 

While our two-language setup successfully isolates the pairwise mechanisms of cross-lingual transfer, real-world multilinguality is inherently combinatorial. Extending this analysis to multiple languages is an important future direction, though it requires modeling complex asymmetries, such as typological pivoting or interference across language families.

\textbf{Limitations}
Our synthetic languages are defined solely as sets of templates, ignoring structural and lexical (dis)similarities between languages. Also, the attributes are spelled the same across our synthetic languages which is true for some attribute types (e.g., years or numbers) and some language pairs (e.g., Paris is spelled the same in several languages). While these are clear simplifications of LLM pretraining data, our key findings are independent of these design choices.  
Finally, our experiments focus exclusively on single-hop atomic recall (mapping a subject and relation to an attribute). They don't indicate whether the unification mechanism extends to complex, compositional, or multi-hop factual reasoning, which represents a distinct and harder regime \citep{biran-2024-hopping,wang-2024-grokking}.

\section{Conclusions}

In this work we use a controlled setting to study why LMs often fail at cross-lingual knowledge transfer, hallucinating facts in one language that they know in another. We demonstrate how these failures are caused by models developing separate, language-specific representations for facts rather than unified, language-agnostic ones. Our key finding is that separation is driven by the \textit{informativeness} and \textit{extractability} of the language feature itself and happens very early in training. 
These results shed light on the role of tokenization, shared vocabulary, script and embedding alignment which have been observed previously but without explanation.
Finally, we introduce a language unification score to quantify the representational similarity which is both strongly predictive of cross-lingual factual accuracy and useful for checkpoint selection.

\section*{Acknowledgements}

We thank Andrew Lampinen, Michael Lepori, Chris Dyer, Nithum Thain and Alicia Machado for the very helpful feedback. 

\bibliography{colm2026_conference}
\bibliographystyle{colm2026_conference.bst}

\newpage

\section{Appendix A: Additional Details}\label{sec:appendix}

\subsection{Data generation procedure}\label{app:data-gen}

\begin{figure*}[t]
  \centering
  \begin{subfigure}{0.49\textwidth}
    \centering
    \includegraphics[width=0.98\linewidth]{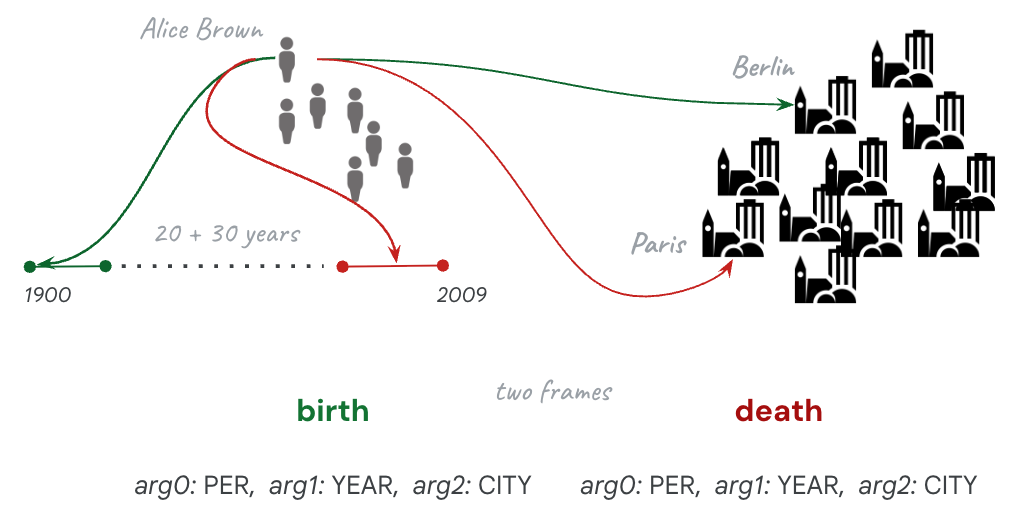}
    \caption{Knowledge Graph (KG) is populated with events of two types}
    \label{fig:data:events}
  \end{subfigure}
  \hfill
  \begin{subfigure}{0.49\textwidth}
    \centering
    \includegraphics[width=0.95\linewidth]{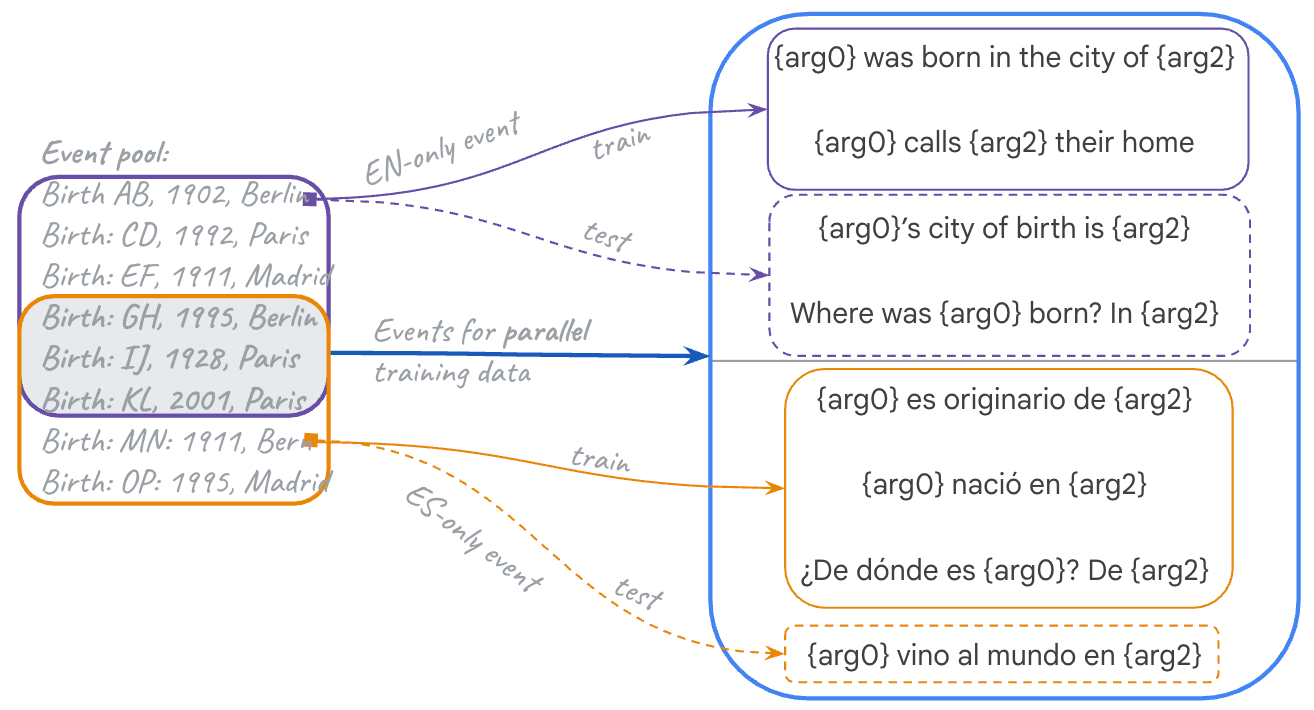}
    \caption{Templates to express birth-city facts in EN and ES}
    \label{fig:data:templates}
  \end{subfigure}

  \caption{\textcolor{green}{Birth} and \textcolor{purple}{death} events are created for every entity by sampling from a set of years (disjoint) and cities (same pool). A dataset is comprised of monolingual (expressed in either \textcolor{violet}{EN} or \textcolor{orange}{ES}) and parallel (expressed in \textcolor{blue}{both} languages) events. Arrows point from a particular event to the training templates (solid line) or in-language test templates (dashed). All verbalizations in the other language are part of the cross-lingual test set. To simplify, only birth templates with \textit{arg0, arg2} are shown.}
  \label{fig:data}
  \vspace{-1em}
\end{figure*}

For readability, Fig.~\ref{fig:data} uses English and Spanish templates, but in our setup templates are constructed by randomly sampling tokens from a predefined vocabulary (see App.~\ref{app:templates} for examples). No two templates share tokens, so no tokens are shared
between languages (except for arguments). In all experiments we use \textbf{two languages}, mostly with five templates per language and fact type.
See pseudo-code for the data generation process below; sample templates are shown in App.~\ref{app:templates}.
Since cross-lingual events are seen in (at least) twice as many training examples as the events which are only expressed in a single language, we need to ensure that the total size of the training dataset is invariant to the fraction of parallel data so we can compare models fairly. To this end, we downsample verbalizations of monolingual events. 
In our experiments the attributes are sampled uniformly, which naturally results in a variability in attribute frequencies (e.g., there may be more people born in \textit{Berlin} than in \textit{Paris}). We have a set of 100 cities (joint between the birth and death events), and two disjoint sets of 20 \& 30 years for birth and death years. 
We also create a custom tokenizer which tokenizes every word as a single token. We include pseudocode for the knowledge-graph generation and partitioning in Fig. \ref{fig:kg-pseudocode}.

\begin{figure}
\begin{python}
def create_data_splits():
  events = []
  for name in names:
    birth = Event(name, random(cities), random(birth_days))
    death = Event(name, random(cities), random(lifespans) + birth.date)
    events.extend([birth, death])

  templates["lang0"]["birth", "time"] = []
  for _ in range(N_TEMPLATES):
    tokens = random(tokenizer.vocab(), size=(TEMPLATE_LEN,))
    tokens = randomly_insert(" {subject}", tokens)
    tokens.append(" {time}")
    templates["lang0"]["birth", "time"].append(tokens)

  crosslingual_events, other_events = random_split(events)

  crosslingual_train_data = cartesian_product(templates, crosslingual_events)
  other_train_data = {
     lang: cartesian_product(templates[lang], other_events[lang])
     for lang in LANGUAGES
  }

  other_train_data, same_lang_test = drop_equally_by_event(other_train_data)
  if is_too_big(other_train_data):
    other_train_data, extra_same_lang_test = 
        drop_extra_train_data(other_train_data)
    same_lang_test += extra_same_lang_test

  cross_lang_test = {
     lang: cartesian_product(templates - templates[lang], other_events[lang])
     for lang in LANGUAGES
  }
  train_data = other_train_data + crosslingual_train_data
\end{python}
\caption{Pseudocode for data generation \& test set splitting}
\label{fig:kg-pseudocode}
\end{figure}

\subsection{Model \& Training Details}
\label{app:training-details}
The models for most of our experiments use the Gemma-2 architecture with models of approximately 2 million parameters. For most experiments, we use the following parameters:

\begin{center}
\begin{tabular}{l r}
\toprule
\textbf{Hyperparameter} & \textbf{Value} \\
\midrule
Hidden Size ($D_h$) & 128 \\
Intermediate Size ($D_i$) & 512 \\
Head Dimension ($D_{head}$) & 64 \\
\midrule
Number of Hidden Layers ($L$) & 6 \\
Number of Attention Heads ($H_{attn}$) & 4 \\
Number of Key/Value Heads ($H_{kv}$) & 1 \\
\bottomrule
\end{tabular}
\end{center}

We also conduct experiments with Pythia models and with Gemma models with both 0.1x the number of parameters and 10x the number of parameters, but scale did not end up being a substantial variable in this work. Models are trained for 100 epochs (although this is quite gratuitous, they typically converge in a small fraction of this time). We train models with a weight decay of 0.2, although we note that varying this from 0.0 to 0.4 did not change results very much. We train with a batch size of 128 and a cosine learning rate starting at 3e-4.

\begin{figure*}[htbp]
    \centering

    \begin{subfigure}{0.65\linewidth} 
        \centering
        \includegraphics[width=0.9\textwidth]{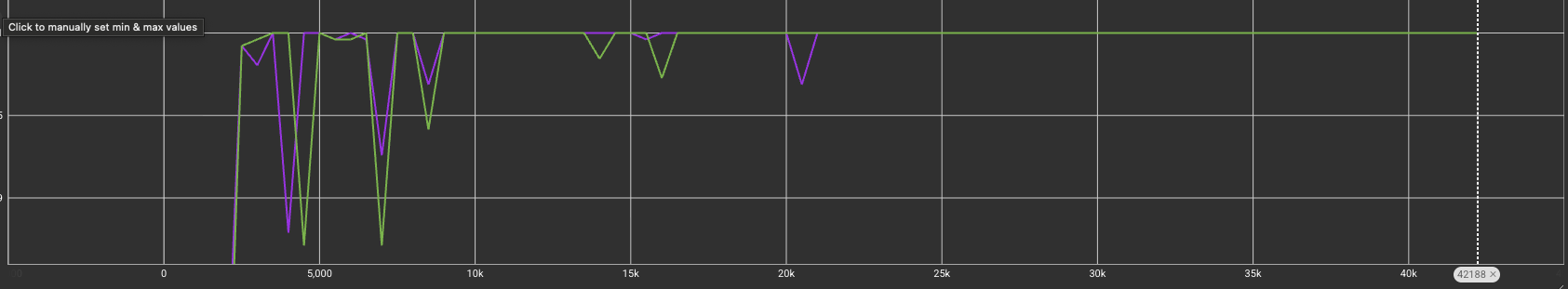} 
        \caption{Accuracy on the training set by step}
        \label{fig:train_performance}
    \end{subfigure}
    
    \vspace{0.5em} 

    \begin{subfigure}{0.65\linewidth}
        \centering
        \includegraphics[width=0.9\textwidth]{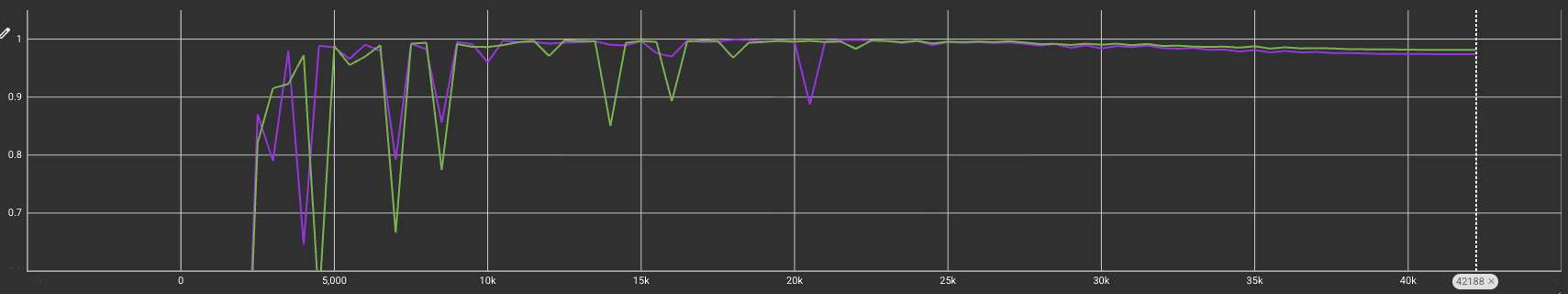}
        \caption{Accuracy on the same-lang test set by step}
        \label{fig:idtest_performance}
    \end{subfigure}

    \vspace{0.5em} 

    \begin{subfigure}{0.65\linewidth}
        \centering
        \includegraphics[width=0.9\textwidth]{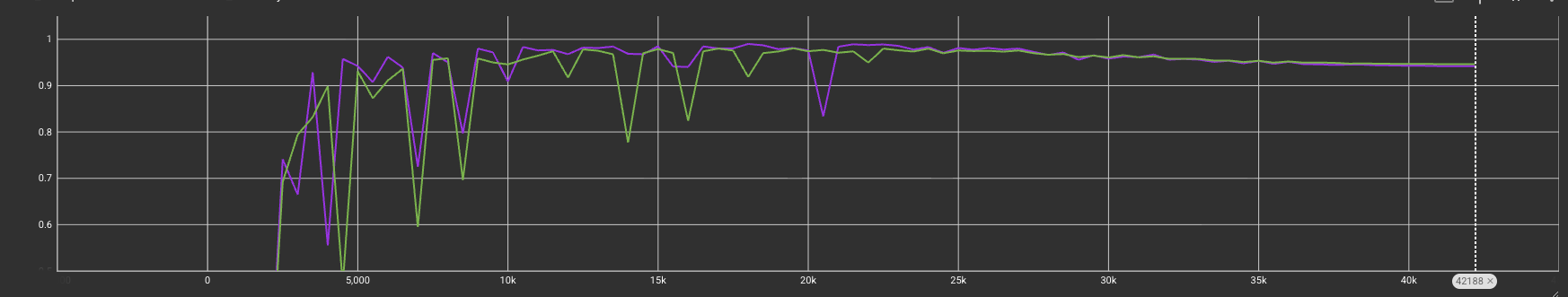}
        \caption{Accuracy on the cross-lang test set by step}
        \label{fig:oodtest_performance}
    \end{subfigure}

    \caption{Model performance across various datasets. Subfigures (a), (b), and (c) display results for the training, same language test, and cross-lingual test sets, respectively.}
    \label{fig:model_performance}
\end{figure*}

We include reference training plots in \ref{fig:model_performance}.


\subsection{Sample Dataset}\label{app:templates}
Every time we create a dataset, we generate two new "languages" made of new templates. We show here all the templates for one of these datasets and a fragment of the resulting dataset.

  \resizebox{0.99\linewidth}{!}{%
  \begin{tabular}{c c l}
  \hline
  \textbf{Language ID} & \textbf{Frame} & \textbf{Template} \\
  \hline
  0 & birth & \{arg0\} Ramoss recib 58 citizenship An 1965 verlassen \{arg1\} \\
  0 & birth & 1922 cel 117 \{arg0\} top Barcel \{arg2\} \\
  0 & birth & \{arg0\} died primer Rob Olivers Leuv ist 93 \{arg1\} \\
  0 & birth & \{arg0\} Johnstons mar Youn German New \{arg2\} \\
  0 & birth & ran hails \{arg0\} hap Hansens by \{arg1\} \\
  0 & birth & deceas 107 read Montgom \{arg0\} Per \{arg2\} \\
  0 & birth & p 65 Col 1967 \{arg0\} Hendersons Grah 57 \{arg1\} \\
  0 & birth & \{arg0\} Morgensterns born 2012 Leice welcomed \{arg2\} \\
  0 & birth & \{arg0\} 104 Chapmans Whose tuv Mur \{arg1\} \\
  0 & birth & di \{arg0\} Wrights major Hawkinss Klagen \{arg2\} \\
  0 & death & \{arg0\} 108 encontró origins Frank 99 \{arg1\} \\
  0 & death & 64 1944 next or ta \{arg0\} c \{arg2\} \\
  0 & death & \{arg0\} Blacks Andersons 2018 1960 Phoen \{arg1\} \\
  0 & death & \{arg0\} Lugan destination descans Coopers Podolskis Kelleys Duncans \{arg2\} \\
  0 & death & \{arg0\} Angel 16 Bishops 1951 Winnip Where 1927 \{arg1\} \\
  0 & death & Danielss Nant concerts Tur \{arg0\} 1933 nati \{arg2\} \\
  0 & death & music Rey Adam Perrys of \{arg0\} descanso fal \{arg1\} \\
  0 & death & Philadelph Thomass \{arg0\} Chavezs 19 Woodss \{arg2\} \\
  0 & death & Diazs Par 62 1969 \{arg0\} Miss Manche \{arg1\} \\
  0 & death & \{arg0\} 200 Johnstons 6 Who Meyers Harpers \{arg2\} \\
  \hline
  1 & birth & \{arg0\} Montgom verstarb Diazs Foxs Blacks mundo 1949 \{arg1\} \\
  1 & birth & \{arg0\} hat Aust meet V from Wh \{arg2\} \\
  1 & birth & \{arg0\} 3 Watkinss Hend 21 Bankss Thomass Ott \{arg1\} \\
  1 & birth & Svet Morgans An \{arg0\} Lawrences Mart tied \{arg2\} \\
  1 & birth & Jamess Mont Br Ar raised \{arg0\} Daviss \{arg1\} \\
  1 & birth & \{arg0\} 85 Gardners Rices Wat 70 83 \{arg2\} \\
  1 & birth & Sanderss \{arg0\} 81 65 48 fallecimiento Mey \{arg1\} \\
  1 & birth & Scotts 61 pass Ed \{arg0\} Rosss \{arg2\} \\
  1 & birth & \{arg0\} studen Vasqu 40 Tuckers many erblickte 87 \{arg1\} \\
  1 & birth & \{arg0\} Hills 13 1971 Petersons 1951 encontró \{arg2\} \\
  1 & death & Torress es Camp married mund \{arg0\} 45 26 \{arg1\} \\
  1 & death & \{arg0\} seine Angeles 98 cam im \{arg2\} \\
  1 & death & Hendersons \{arg0\} Sander tuv reader V raised \{arg1\} \\
  1 & death & \{arg0\} Po Jordans Dorn 76 Cal Whit Brow \{arg2\} \\
  1 & death & 2019 Longs 28 Knig listen \{arg0\} orig \{arg1\} \\
  1 & death & Gre Vanc \{arg0\} finali Ess Griff listening 42 \{arg2\} \\
  1 & death & chil k Tay Carters \{arg0\} Fiel recib When \{arg1\} \\
  1 & death & Geburts fallec dece \{arg0\} Lich Ortizs Parkers 35 \{arg2\} \\
  1 & death & \{arg0\} verlass Boy likes Co Svet Tou \{arg1\} \\
  1 & death & \{arg0\} Sc Martinezs nati Colemans 113 Mc \{arg2\} \\
  \hline
  \end{tabular}%
}
  
This is then an excerpt of six frames for this dataset. This dataset alone contains 2000 frames.

  \begin{lstlisting}[basicstyle=\ttfamily\small, breaklines=true]
  {
    "frame": {
      "core": "birth",
      "args": {"arg0": "PERSON", "arg1": "TIME", "arg2": "CITY"}
    },
    "arg_values": {"arg0": "Aaliyah Anderson", "arg1": "1906", "arg2": "Sheffield"}
  }
  {
    "frame": {
      "core": "death",
      "args": {"arg0": "PERSON", "arg1": "TIME", "arg2": "CITY"}
    },
    "arg_values": {"arg0": "Aaliyah Anderson", "arg1": "1990", "arg2": "Zurich"}
  }
  {
    "frame": {
      "core": "birth",
      "args": {"arg0": "PERSON", "arg1": "TIME", "arg2": "CITY"}
    },
    "arg_values": {"arg0": "Aubrey Griffin", "arg1": "1914", "arg2": "Portland"}
  }
  {
    "frame": {
      "core": "death",
      "args": {"arg0": "PERSON", "arg1": "TIME", "arg2": "CITY"}
    },
    "arg_values": {"arg0": "Aubrey Griffin", "arg1": "1996", "arg2": "Innsbruck"}
  }
  {
    "frame": {
      "core": "birth",
      "args": {"arg0": "PERSON", "arg1": "TIME", "arg2": "CITY"}
    },
    "arg_values": {"arg0": "Luke Dunn", "arg1": "1917", "arg2": "Edmonton"}
  }
  {
    "frame": {
      "core": "death",
      "args": {"arg0": "PERSON", "arg1": "TIME", "arg2": "CITY"}
    },
    "arg_values": {"arg0": "Luke Dunn", "arg1": "1997", "arg2": "Rostov"}
  }
  \end{lstlisting}
  
Finally, we create our datasets by expressing the events contained in these frames on a subset of templates. Here is a fragment of the resulting dataset: 

\resizebox{0.99\linewidth}{!}{%
  \begin{tabular}{c c l l}
  \hline
  \textbf{Language} & \textbf{Frame} & \textbf{Person} & \textbf{Generated Text} \\
  \hline
  0 & birth & Aaliyah Anderson & Aaliyah Anderson Ramoss recib 58 citizenship An 1965 verlassen 1906 \\
  0 & birth & Aaliyah Anderson & 1922 cel 117 Aaliyah Anderson top Barcel Sheffield \\
  1 & birth & Aaliyah Anderson & Aaliyah Anderson Montgom verstarb Diazs Foxs Blacks mundo 1949 1906 \\
  1 & birth & Aaliyah Anderson & Aaliyah Anderson hat Aust meet V from Wh Sheffield \\
  0 & birth & Aubrey Griffin & Aubrey Griffin Ramoss recib 58 citizenship An 1965 verlassen 1914 \\
  0 & birth & Aubrey Griffin & 1922 cel 117 Aubrey Griffin top Barcel Portland \\
  1 & birth & Luke Dunn & Luke Dunn Montgom verstarb Diazs Foxs Blacks mundo 1949 1917 \\
  1 & birth & Luke Dunn & Luke Dunn hat Aust meet V from Wh Edmonton \\
  0 & death & Aaliyah Anderson & Aaliyah Anderson 108 encontró origins Frank 99 1990 \\
  0 & death & Aaliyah Anderson & 64 1944 next or ta Aaliyah Anderson c Zurich \\
  1 & death & Aaliyah Anderson & Torress es Camp married mund Aaliyah Anderson 45 26 1990 \\
  1 & death & Aaliyah Anderson & Aaliyah Anderson seine Angeles 98 cam im Zurich \\
  0 & death & Aubrey Griffin & Aubrey Griffin 108 encontró origins Frank 99 1996 \\
  0 & death & Aubrey Griffin & 64 1944 next or ta Aubrey Griffin c Innsbruck \\
  1 & death & Luke Dunn & Torress es Camp married mund Luke Dunn 45 26 1997 \\
  1 & death & Luke Dunn & Luke Dunn seine Angeles 98 cam im Rostov \\
  \hline
  \end{tabular}%
}

\subsection{Unification+ Score For Model Selection}\label{app:unif-score}
At random initialization, the language unification score is typically around 1.0, because there is no more variation across-languages than there is within language. 
As a result of this, we utilize a slightly modified version of the language unification score when selecting the best checkpoint. For this, we multiply the language unification score with the same-lang test accuracy. Intuitively, this balances a model that fits the existing data well with one that generalizes well. This metric is listed as `unification-score+` in the chart.

\begin{figure}[!h]
\begin{python}
def compute_unification_scores(model, datapoints):
  total = 0
  for datapoints_by_fact in group_by_fact(datapoints):
    for idx in range(len(datapoints_by_fact)):
      datum = datapoints_by_fact[idx]
      same_lang = [
        d 
        for i, d in enumerate(datapoints_by_fact) 
        if d.language == datum.language and i != idx
      ]
      other_lang = [
        d 
        for i, d in enumerate(datapoints_by_fact) 
        if d.language == datum.language and i != idx
      ]
      total += mean_cosine_similarity(model, datum, other_lang) / 
        mean_cosine_similarity(model, datum, same_lang)

  return total / len(datapoints)

def mean_cosine_similarity(model, datum, other_data):
  return mean([
    cosine_similarity(
      concat_residual_latents_at_last_token(model, datum),
      concat_residual_latents_at_last_token(model, other_datum)
    )
    for other_datum in other_data
  ])
\end{python}
\caption{Pseudocode for calculation of language unification scores}
\label{fig:unif-score-pseudocode}
\end{figure}

\section{Appendix B: Complementary Results}
\subsection{Gradient-based representations yield similar patterns}
\begin{figure}[H]
\centering
    \includegraphics[width=0.45\linewidth]{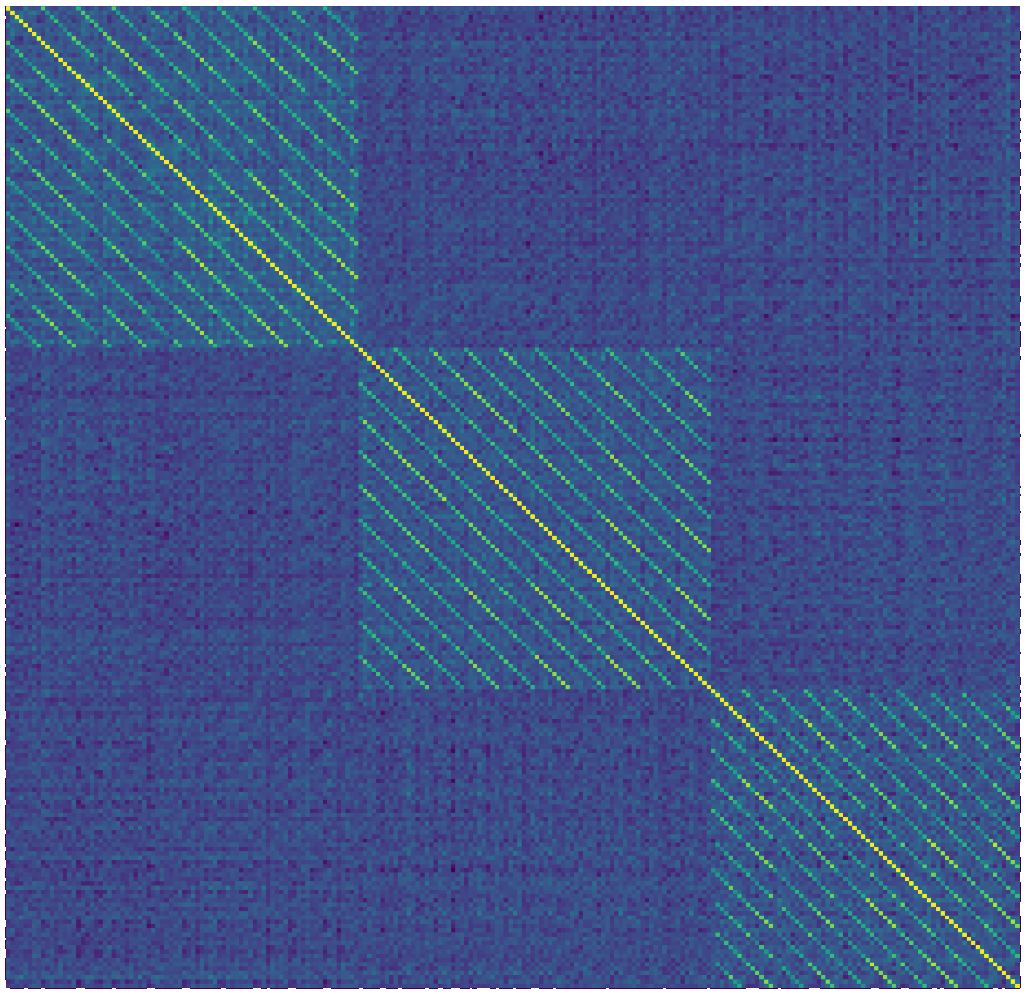}
    \hfill
    \includegraphics[width=0.45\linewidth]{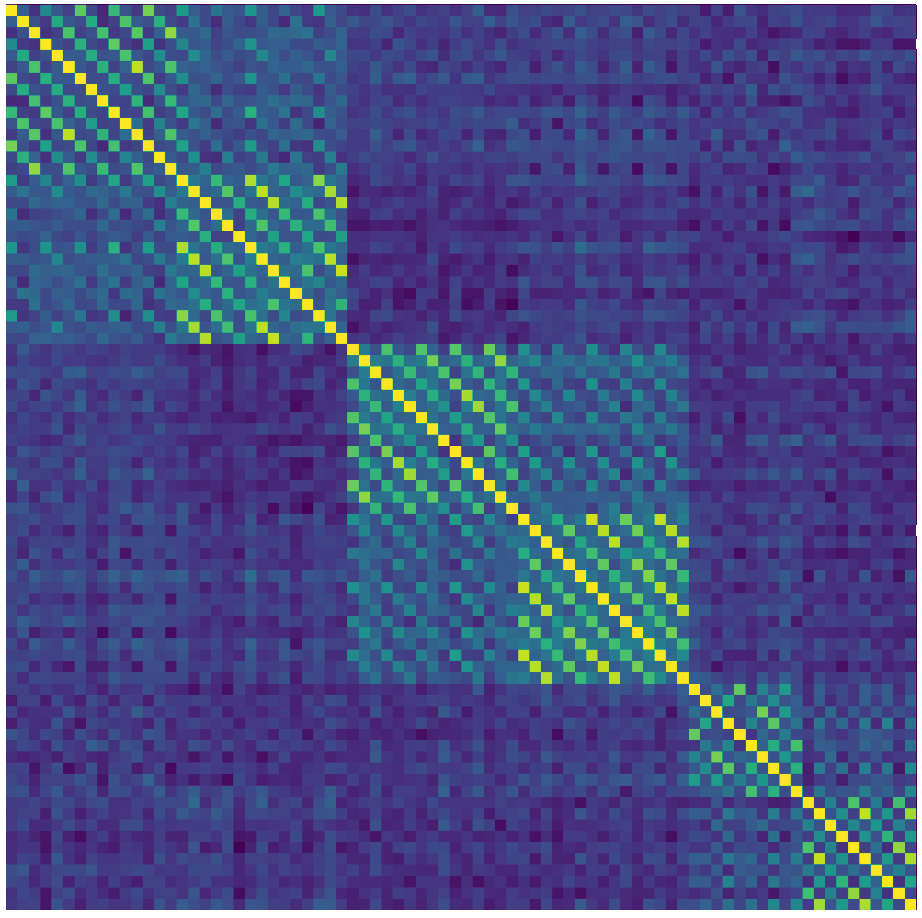}
    \caption{[Models with 30\% and 4\% cross-lingual events, final checkpoint] Also with \textbf{gradient-based} representations the language checkerboarding is visible in a model with poor cross-lingual generalization (right). Three most frequent (in the cross-lingual portion of the respective training data) birth-city attributes are picked to collect representations. }
  \label{fig:app:plaids-ex:gradient}
\end{figure}

\subsection{Layerwise results for activation language unification scores vs xLang Test Acc}

\begin{figure}[H]
    \centering
    \includegraphics[width=0.95\linewidth]{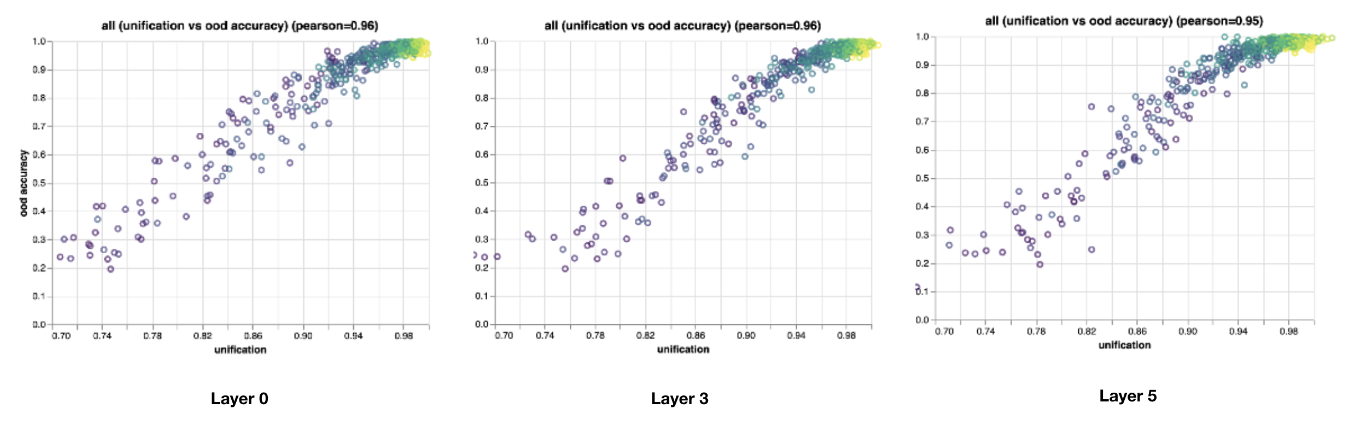}
    \caption{Layerwise results for activation language unification scores vs. xLang Test Acc.}
    \label{fig:activ-ood-layerwise}
\end{figure}

\subsection{Language checkerboarding is more obvious for lower layers}
\begin{figure}[H]
\centering
    \includegraphics[width=0.45\linewidth]{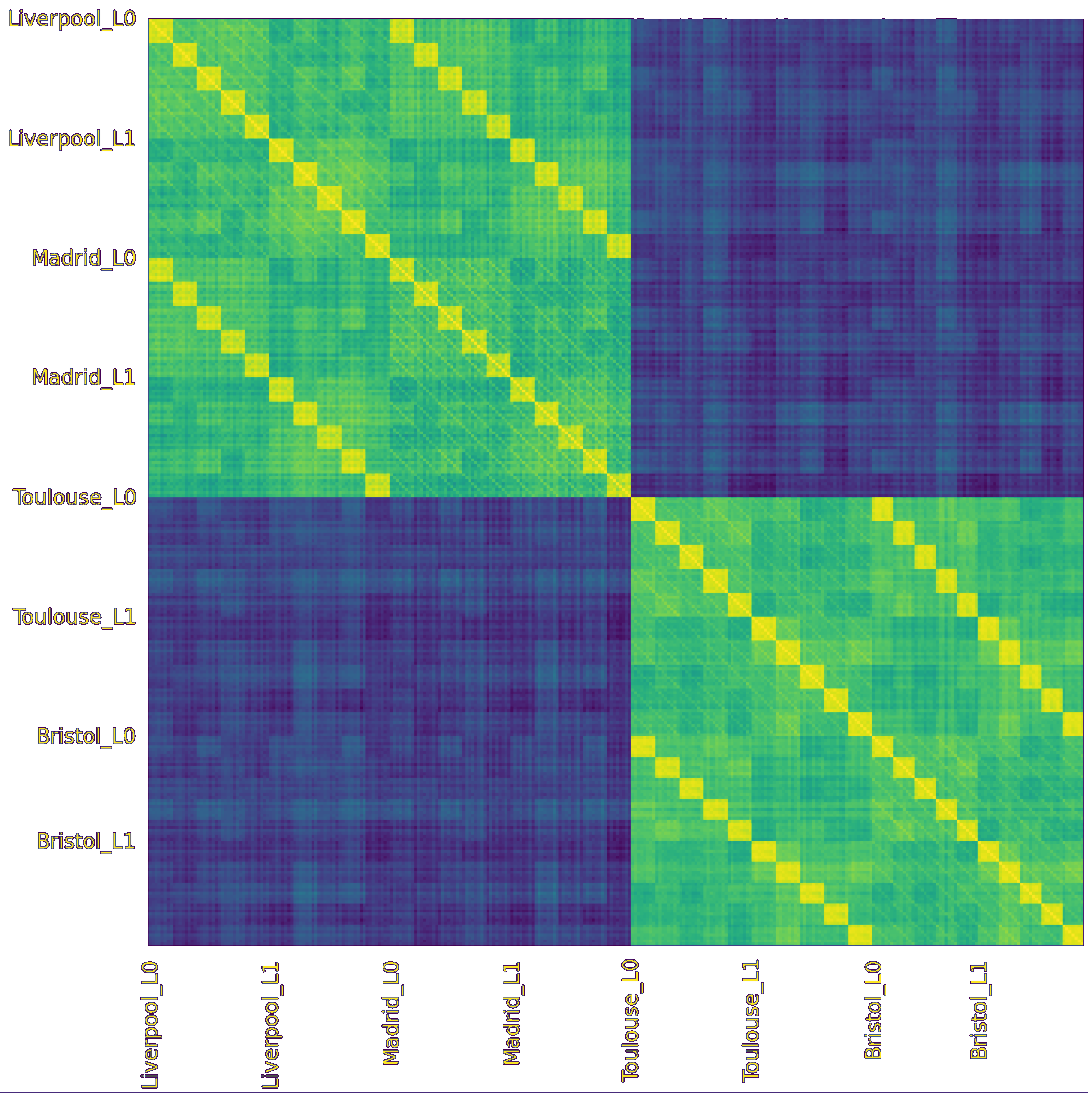}
    \hfill
    \includegraphics[width=0.45\linewidth]{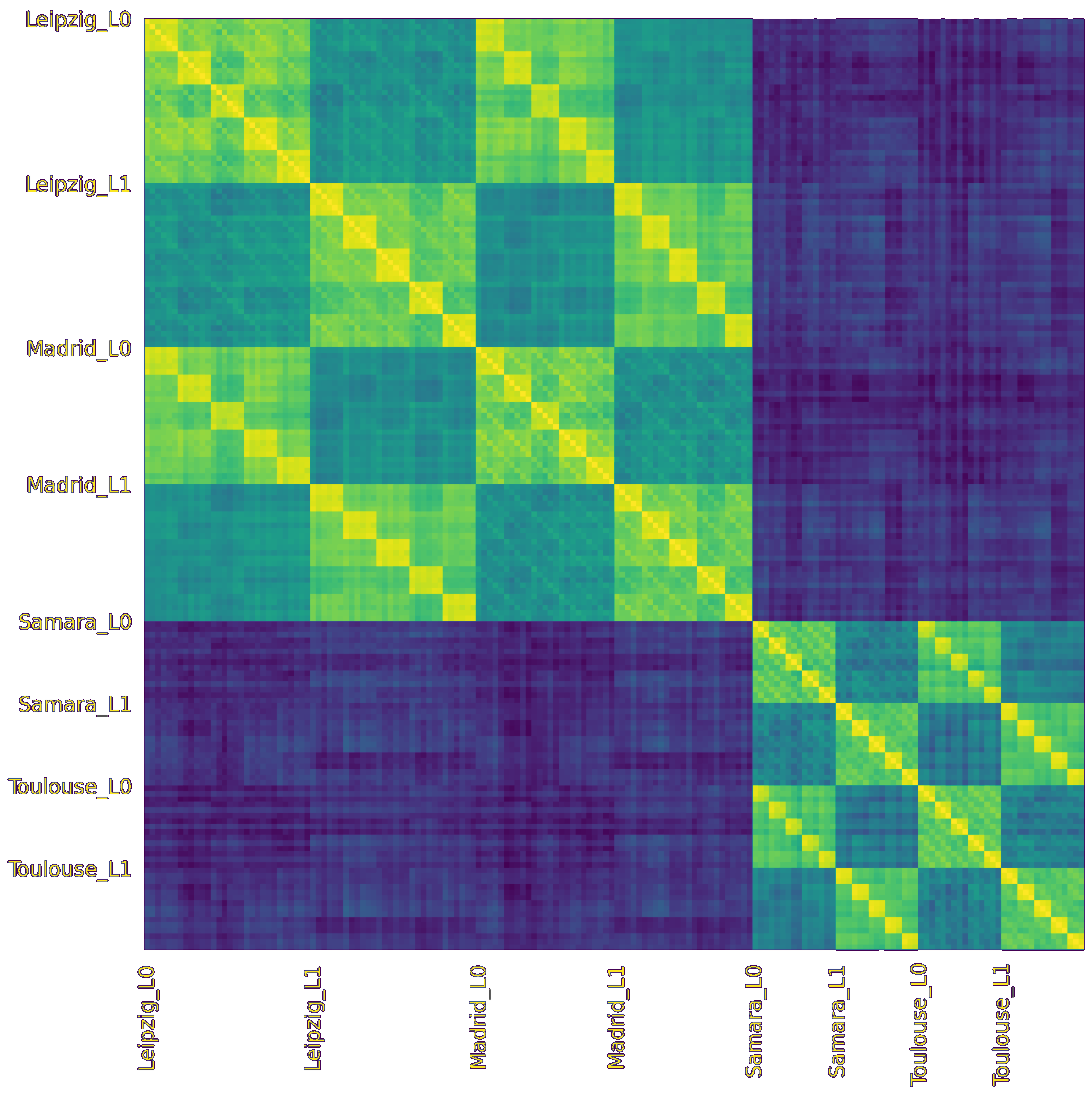}
    \caption{[Models with 30\% and 8\% cross-lingual events, layers 0-1, final checkpoint, activations-based representations] The presence of language checkerboarding (right) is particularly striking in the lowest model layers. For the poorly generalizing model, language identity is the dominant factor determining the similarity of representations for the same fact type. Two birth-city and two death-city attributes (most frequent in the cross-lingual portion of the respective training data) are picked to collect representations.}
  \label{fig:app:plaids-ex:layers0-1}
\end{figure}

\subsection{Language checkerboarding is more obvious for imbalanced models}

\subsection{Language checkerboarding is more obvious for cased models}

\subsection{Adding templates improves cross-lingual generalization}
\label{app:increasing-templates}
\begin{figure}[H]
    \centering
    \includegraphics[width=0.7\linewidth]{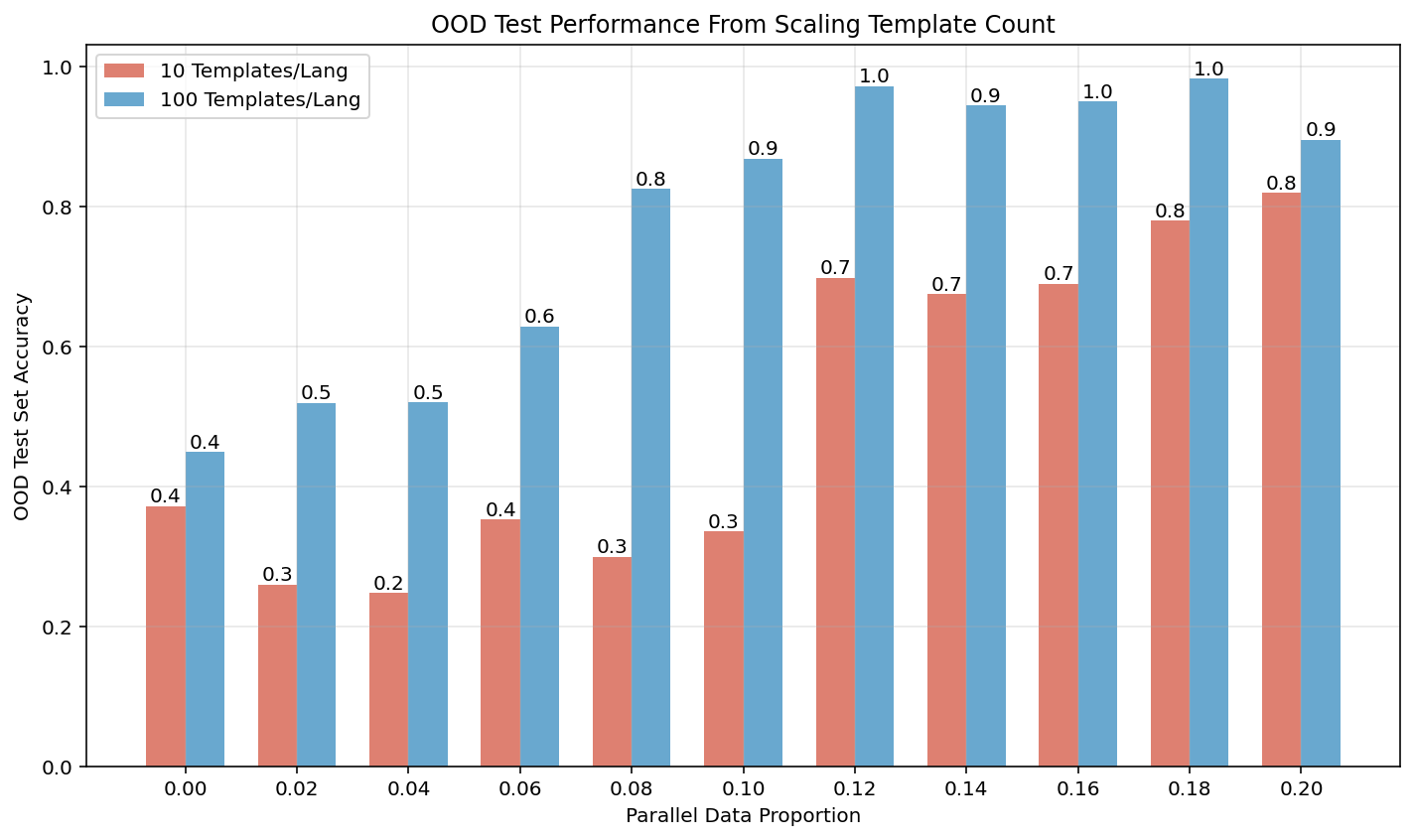}
    \caption{Increasing the number of templates substantially improves cross-lingual generalization, despite also increasing the difficulty of the test set.}
    \label{fig:more-templates}
\end{figure}

\subsection{Vocab Similarity vs Cross-Lingual Accuracy}\label{app:vocab-sim}

\begin{figure}[H]
\centering
\centering
    \includegraphics[width=0.45\linewidth]{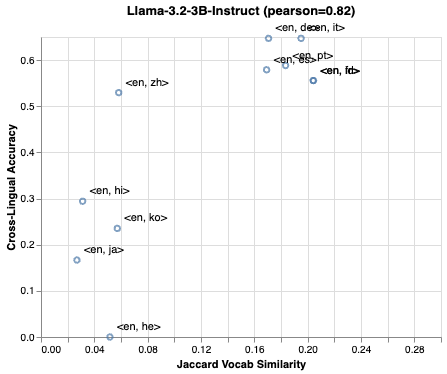}
    \includegraphics[width=0.45\linewidth]{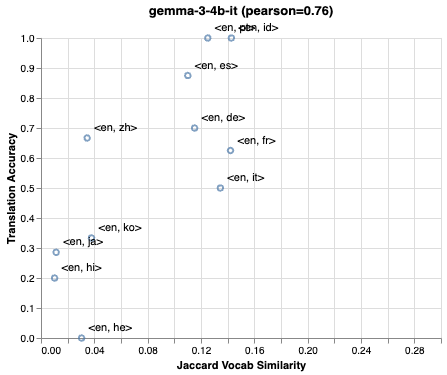}
    \includegraphics[width=0.45\linewidth]{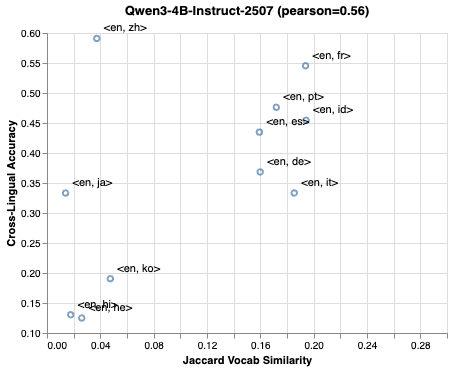}
    \includegraphics[width=0.45\linewidth]{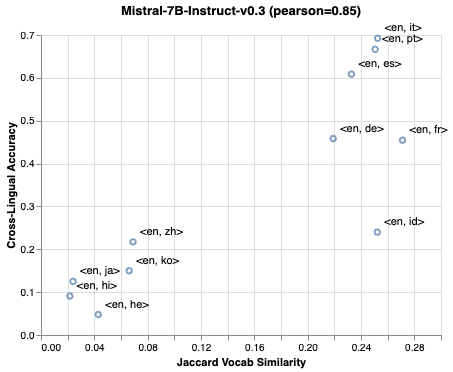}
    \caption{Cross-Lingual accuracy vs. vocabulary similarity. $<x,y>$ indicates going from source language $x$ to target language $y$. For most models, the lower left includes languages like Hebrew that have little to no vocabulary overlap with English. On the other hand, languages such as Indonesian tend to have higher vocabulary overlap and higher cross-lingual accuracy. The average correlation coefficient is 0.69, ranging from 0.43 to 0.85 across different models.}\label{fig:results:llm-source-eng}

\end{figure}

\begin{table}[H]
\caption{Percentage of variance explained in cross-lingual accuracy when considering vocabulary overlap between source and target languages.}
\label{tab:vocab-overlap}
\centering
\resizebox{\linewidth}{!}{%
\begin{tabular}{l|cccccccccccc}
\hline
\textbf{Source Lang.} & \textbf{de} & \textbf{en} & \textbf{es} & \textbf{fr} & \textbf{he} & \textbf{hi} & \textbf{id} & \textbf{it} & \textbf{ja} & \textbf{ko} & \textbf{pt} & \textbf{zh} \\ \hline
\textbf{\# Datapoints} & 90 & 265 & 95 & 45 & 80 & 210 & 90 & 120 & 140 & 65 & 95 & 85 \\
\textbf{$R^2$ (\%)} & 68.93 & 32.75 & 69.06 & 99.53 & 31.30 & 41.52 & 37.60 & 79.99 & 0.03 & 45.94 & 24.14 & 0.20 \\ \hline
\end{tabular}
}
\end{table}

\subsection{Fact ID vs language ID feature footprints.}
\label{app:facts-vs-lang-footprint}

\begin{figure}[H]
    \centering
    \includegraphics[width=1.0\linewidth]{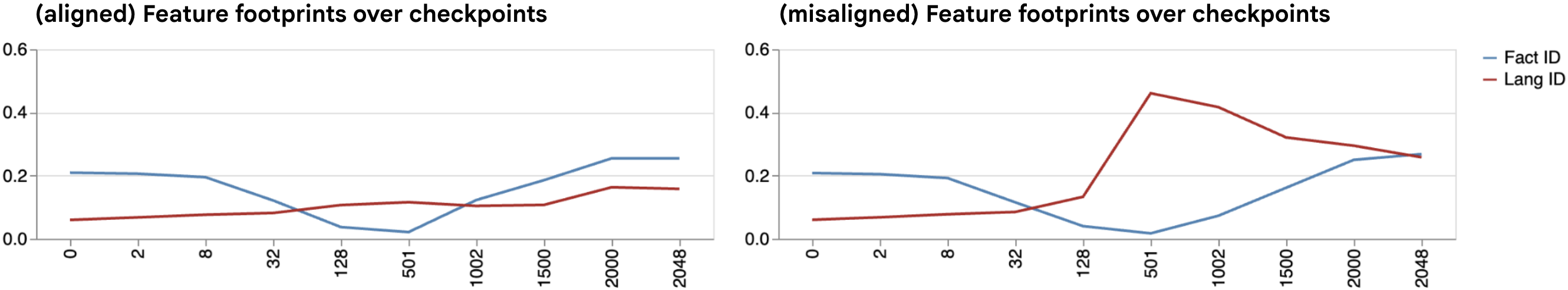}
    \caption{Representational variance explained by language (``Lang ID'' - red) vs. fact (``Fact ID'' - blue) across initial checkpoints. Language is uninformative for the model on the left (\texttt{Balanced}), whereas it is highly informative for the model on the right (\texttt{Imbalanced}). Both models are trained with 0\% parallel data, thus language identity is highly extractable. However in the \texttt{Imbalanced} model, language identity is also highly informative. The language feature footprint grows quickly in the imbalanced model (before shrinking, although it still ends up larger in imbalanced than balanced models - see Fig. \ref{fig:results:aligned}).}
    \label{fig:facts_vs_lang_footprints}
\end{figure}

\end{document}